\documentclass{article}

\usepackage{tabularx}
\usepackage{booktabs}    % for professional tables
\usepackage{graphicx}
\usepackage{multirow}
\usepackage{subcaption}
\usepackage{xcolor}
\usepackage{dsfont}
\usepackage{wrapfig}
\usepackage{bm}
\usepackage{amsmath}
\usepackage{amssymb}
\usepackage{microtype}
\usepackage[preprint]{corl_2026} % Uncomment for pre-prints (e.g., arxiv); This is like ``final'', but will remove the CORL footnote.

\newcolumntype{Y}{>{\centering\arraybackslash}X}
\newcommand{\paperTable}{%
    \scriptsize%
    \setlength{\tabcolsep}{3pt}%
    \renewcommand{\arraystretch}{1.0}%
}
\newcommand{\paperTableTight}{%
    \scriptsize%
    \setlength{\tabcolsep}{1.5pt}%
    \renewcommand{\arraystretch}{1.0}%
}

\title{HiMoE-VLA:\! Hierarchical\! Mixture-of-Experts for Generalist Vision-Language-Action Policies}

\author{Zhiying Du$^{1}$\thanks{Work done during research internship at Microsoft Research.}, \
  Bei Liu$^{2}$, \
  Yaobo Liang$^{2}$, \
  Yichao Shen$^{3\ast}$, \
  Haidong Cao$^{1}$, \
  Xiangyu Zheng$^{1}$, \\
  \textbf{Zhiyuan Feng}$^{4\ast}$, \
  \textbf{Zuxuan Wu}$^{1}$\thanks{Corresponding author.}, \
  \textbf{Jiaolong Yang}$^{2\dagger}$, \
  \textbf{Yu-Gang Jiang}$^{1}$ \\
  \\
  $^{1}$Fudan University \quad
  $^{2}$Microsoft Research Asia \quad \\
  $^{3}$Xi’an Jiaotong University \quad
  $^{4}$Tsinghua University \\
  \\
}

\begin{document}
\maketitle

%===============================================================================

\begin{abstract}
Generalist vision--language--action (VLA) policies are typically trained on heterogeneous mixtures of robot demonstrations spanning diverse embodiments, action spaces, and observation configurations. Modeling such heterogeneity with a shared dense action module can induce negative transfer, particularly when action spaces or visual observations differ across data sources. We address this issue with HiMoE-VLA, a VLA framework built around a Hierarchical Mixture-of-Experts (HiMoE) action module. HiMoE uses Action-Space MoE layers at the input/output boundaries to specialize computation for distinct action spaces, Heterogeneity-Balancing MoE layers in neighboring layers to provide balanced capacity for residual variation in observations, scenes, and embodiments, and dense Transformer blocks in the middle to integrate shared representations. Two auxiliary objectives further guide this hierarchy: a contrastive Action-Space Regularization objective for boundary specialization and a load-balancing objective for stable expert utilization. HiMoE-VLA reaches 3.98 on CALVIN, 98.0\% on LIBERO, and 75.0\% and 63.7\% average success on real xArm7 and ALOHA tasks; under controlled heterogeneous co-training, it turns the negative transfer observed in strong baselines into positive transfer.  The code and models are publicly available at \url{https://github.com/ZhiyingDu/HiMoE-VLA}.
\end{abstract}


%===============================================================================

\section{Introduction}
Vision--language models (VLMs) have emerged as powerful multimodal representation learners~\citep{beyer2024paligemma,touvron2023llama,achiam2023gpt,jiang2023clip,yang2025qwen3,xiao2024florence}, motivating their use as backbones for vision--language--action (VLA) policies that map visual observations and language instructions to robot actions~\citep{brohan2022rt,stone2023open,reuss2025flower,gr3,wang2025unified,xvla,bjorck2025gr00t,shi2025memoryvla,zhou2025chatvla,shukor2025smolvla}. Scaling such policies, however, differs fundamentally from scaling vision or language models: robot data are not organized around a single standardized input--output interface. Demonstrations vary across embodiments, action spaces, state representations, control frequencies, camera viewpoints, and teleoperation or collection protocols. These variations are not merely nuisance factors; they alter the semantics of both observations and action supervision, making naive data mixing prone to interference and limiting transfer across datasets and embodiments.

Recent VLA models~\citep{zitkovich2023rt2,kim2024openvla,team2024octo,black2024pi_0,li2024cogact,spatialvla,openvla-oft,liu2024rdt} commonly pre-train on large heterogeneous mixtures such as Open X-Embodiment (OXE)~\citep{o2024open} and then fine-tune on target domains. This recipe is effective, but most models still process heterogeneous action and observation signals through a shared dense action module. As we show in controlled co-training studies (Tables~\ref{tab:hetero-cotraining} and~\ref{tab:sensor-hetero}), this design can turn additional heterogeneous data into \emph{negative transfer} when action spaces or observation configurations differ. The central question is therefore how to structure the action module so that incompatible factors are separated while transferable structure remains shared.

To address this problem, we propose \textbf{HiMoE-VLA}, a VLA framework with a \textbf{Hierarchical Mixture-of-Experts (HiMoE)} action module. HiMoE assigns different sources of robot-data heterogeneity to different depths of the action module. Because data from different action spaces carry distinct physical semantics and are largely non-transferable across parameterizations (Appendix~\ref{appendix:transfer}), HiMoE isolates action-space-specific computation with Action-Space MoE (AS-MoE) layers at the input/output boundaries. Heterogeneity-Balancing MoE (HB-MoE) layers adjacent to those boundaries provide balanced sparse capacity for remaining observation variability, while dense Transformer blocks in the middle integrate the resulting representations into a shared action representation.

HiMoE-VLA is trained with a flow-matching objective for action generation and two routing regularizers matched to these roles. Action-Space Regularization (AS-Reg) applies a supervised contrastive loss to AS-MoE routing distributions, encouraging tokens from the same action-space/embodiment group to use similar expert patterns. Heterogeneity-Balancing Regularization (HB-Reg) adopts the DeepSeekMoE load-balancing loss~\citep{dai2024deepseekmoe} to prevent the HB-MoE expert pool from collapsing onto a small subset of experts. Together, these losses encourage boundary-level specialization while maintaining balanced capacity for remaining heterogeneity.

After pre-training on OXE~\citep{o2024open} and public ALOHA data~\citep{mobile_aloha,aloha_fine_grain,liu2024rdt}, we fine-tune and evaluate HiMoE-VLA on CALVIN~\citep{mees2022calvin}, LIBERO~\citep{liu2023libero}, and two real robot platforms: xArm7 and ALOHA. HiMoE-VLA improves over strong VLA baselines in both simulation and real-world evaluations, and controlled heterogeneity experiments show that it converts the negative transfer of dense baselines into positive transfer under cross-action and observation/scene co-training.

Our contributions are:
\begin{itemize}
    \item We identify action-space heterogeneity as a major source of negative transfer in multi-source VLA training and propose a hierarchical MoE action module that separates action-space-specific computation while retaining balanced capacity for remaining variability.
    \item We introduce AS-Reg and HB-Reg, two routing-level objectives that respectively encourage action-space-specific routing in AS-MoE and balanced expert utilization in HB-MoE.
    \item We validate HiMoE-VLA across CALVIN, LIBERO, xArm7, and ALOHA, and show controlled gains under heterogeneous co-training.
\end{itemize}

%===============================================================================
\section{Related Work}

\paragraph{Vision--Language--Action Policies.}
Recent VLA models couple pretrained language or vision--language backbones with robot action generation, enabling instruction-conditioned manipulation from visual observations. RT-2~\citep{zitkovich2023rt2} and OpenVLA~\citep{kim2024openvla} cast actions as discrete tokens, while RoboFlamingo~\citep{li2023vision}, Octo~\citep{team2024octo}, UniVLA~\citep{univla}, and $\pi_0$~\citep{black2024pi_0} explore continuous or diffusion/flow-based action prediction. Other works exploit video pretraining to learn visuomotor representations from Internet-scale data without explicit action labels~\citep{gr1,gr2,wu2023unleashing}. These methods demonstrate the value of large pretrained backbones for robotics. However, these systems typically use a largely shared action module, leaving action-space and observation heterogeneity to be absorbed by the same dense computation.

\paragraph{Learning from Heterogeneous Robot Data.}
Large multi-robot datasets such as Open X-Embodiment~\citep{o2024open} expose a central difficulty for generalist policies: demonstrations differ not only in tasks, but also in embodiments, state/action parameterizations, observation configurations, and control conventions. Existing methods address this mismatch mainly through interface alignment or conditioning. RDT-1B~\citep{liu2024rdt} unifies state and action representations for bimanual manipulation; HPT~\citep{hpt} uses dataset-specific stems and heads to align diverse inputs and outputs; and recent generalist policies such as SpatialVLA~\citep{spatialvla}, OpenVLA-OFT~\citep{openvla-oft}, and GR00T-style systems~\citep{bjorck2025gr00t} incorporate stronger spatial, action, or embodiment-aware designs. These approaches reduce part of the mismatch, but they do not explicitly separate action-space discrepancies from other sources of robot-data heterogeneity inside the action module. HiMoE-VLA is complementary: it keeps a shared policy backbone and unified interface, while using a hierarchy that isolates action-space-specific computation before allocating balanced capacity to remaining variability.

\paragraph{Mixture of Experts.}
Mixture-of-Experts (MoE) models have been widely used for sparse scaling in LLMs~\citep{LLMMOE1,LLMMOE2}, and later extended to vision~\citep{VideoMOE} and diffusion models~\citep{DITMOE}. Standard MoE layers route each token to a small subset of experts using top-$k$ gating, with auxiliary mechanisms for routing efficiency and load balancing, including hashing-based routing~\citep{roller2021hash}, dynamic expert activation~\citep{guo2024dynamic,wang2024remoe}, and DeepSeek-style balancing losses~\citep{dai2024deepseekmoe}. In contrast, our use of MoE is not primarily a sparse-scaling mechanism: we impose a depth-wise hierarchy in which AS-MoE layers specialize at the action-space boundaries, HB-MoE layers provide balanced capacity for residual heterogeneity nearby, and dense Transformer layers integrate shared representations in the middle. This differs from a flat MoE stack, where all experts at all depths are asked to absorb all sources of variation simultaneously.

\begin{figure*}[t]
    \centering
    \includegraphics[width=\linewidth]{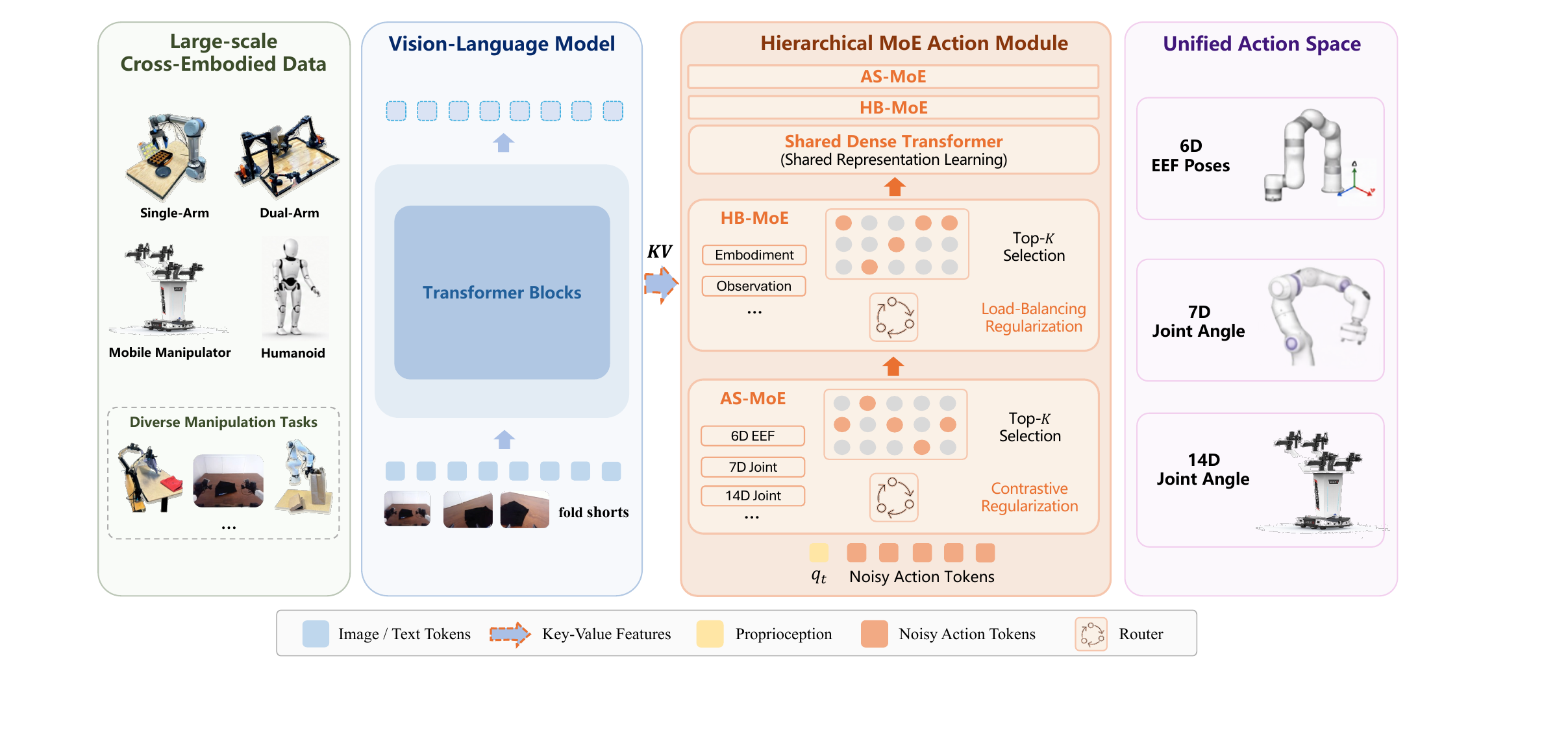}
    \caption{Overview of HiMoE-VLA. The left blue part illustrates the VLM backbone initialized from PaliGemma~\citep{beyer2024paligemma}, and the right orange part depicts our proposed action module with a novel Hierarchical Mixture-of-Experts (HiMoE), which is responsible for processing different robot states and noisy actions and generating final action outputs.}
    \label{fig:overview}
\end{figure*}

%===============================================================================
\section{Method}
\subsection{Problem Formulation}

At time step $t$, the policy receives a language instruction $l$, robot proprioception $q_t$, and RGB observations $o_t=[I_t^1,\ldots,I_t^n]$, and predicts an action chunk
$A_t=[a_t,a_{t+1},\ldots,a_{t+H-1}]$ over horizon $H$:
\[
    \pi_\theta(l,q_t,o_t) \mapsto A_t .
\]
The semantics of $q_t$ and $a_t$ depend on the data source: an action may represent end-effector deltas, joint-angle commands, or their embodiment-specific variants. We therefore map states and actions into a unified vector interface before feeding them to the action module, while retaining a categorical action-space/embodiment identity $c$ for routing regularization. Details of the vector layout, padding, and validity masks are provided in Appendix~\ref{train_details}.

\subsection{Network Architecture}
HiMoE-VLA consists of a pretrained vision--language model and a flow-matching action expert (Fig.~\ref{fig:overview}). The VLM encodes the instruction and multi-view images; the action expert receives proprioception, the noised action chunk, and the flow timestep, and predicts the denoising vector field conditioned on VLM features.

\subsubsection{Vision-Language Module}
We instantiate the VLM with PaliGemma~\citep{beyer2024paligemma}, following $\pi_0$~\citep{black2024pi_0}. Rather than conditioning the action expert only on the final VLM representation, we expose intermediate language-model key--value (KV) states to the corresponding action-expert layers. This layer-wise conditioning lets the action tokens attend to both semantic instruction information and lower-level visual cues; at inference time, the VLM KV states are cached for efficient rollout.
\subsubsection{Action Module with Hierarchical MoE}
\begin{figure*}[t]
    \centering
    \includegraphics[width=0.96\linewidth]{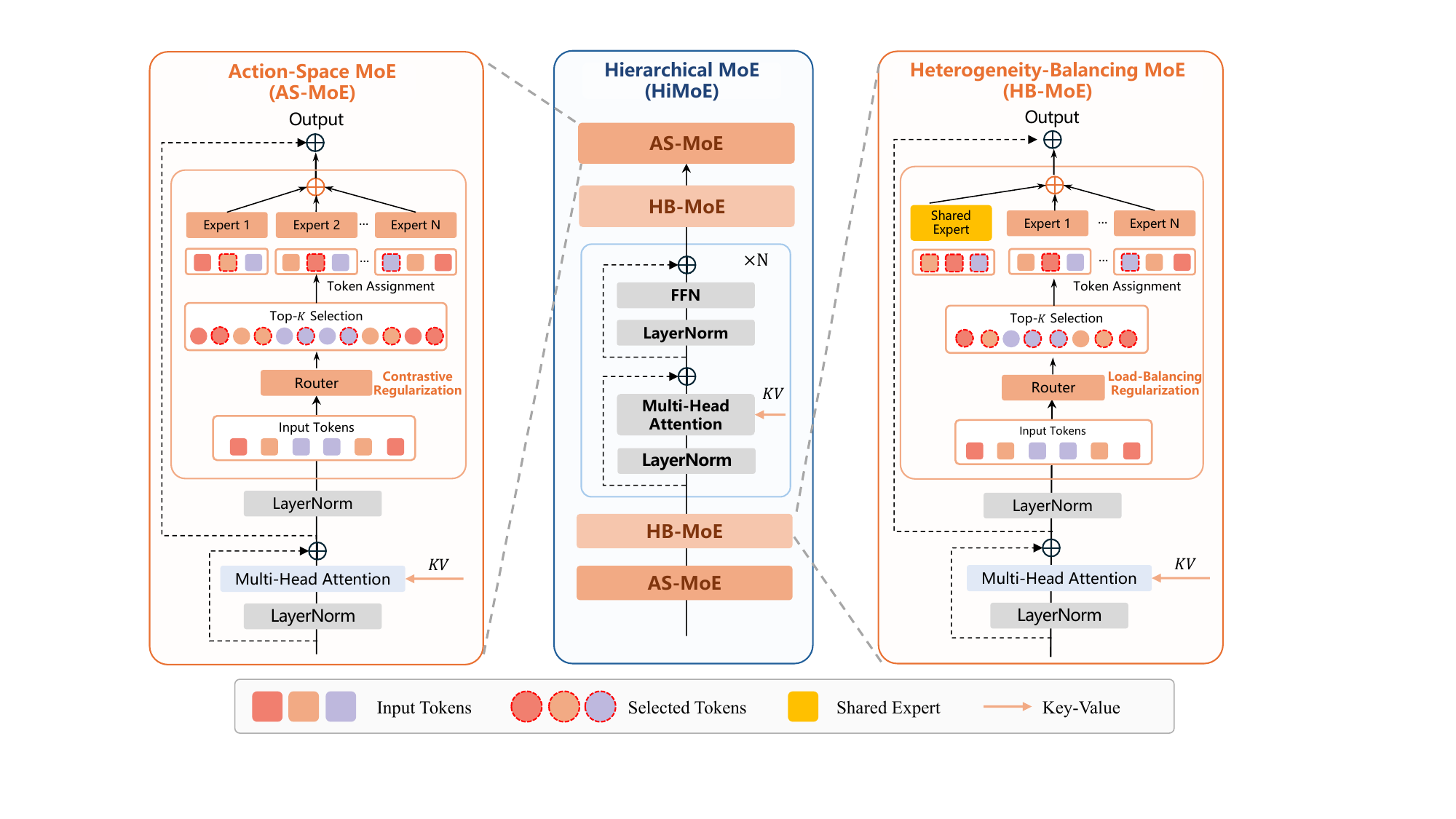}
    \caption{Detailed structure of the Hierarchical Mixture-of-Experts (HiMoE). The architecture follows a layered hierarchy: AS-MoE modules at the boundaries specialize in action-space variations, adjacent HB-MoE modules provide balanced capacity for residual heterogeneity, and the central Transformer blocks serve as shared layers for cross-domain knowledge integration.}
    \label{fig:h_moe_architecture}
\end{figure*}

The action expert embeds the unified proprioceptive vector, the noised action chunk, and the flow timestep into action tokens. It then processes these tokens with a hierarchy of Transformer-style blocks whose feed-forward sublayers are selectively replaced by MoE modules (Fig.~\ref{fig:h_moe_architecture}). The outermost blocks use \textbf{Action-Space MoE (AS-MoE)} to specialize computation for action-space discrepancies such as joint-angle versus end-effector control. The neighboring inner blocks use \textbf{Heterogeneity-Balancing MoE (HB-MoE)} to provide balanced sparse capacity for residual variability such as embodiment and scene differences. The central blocks remain dense Transformers, encouraging information that has passed through the specialization layers to be integrated into a shared representation.

Each MoE block uses top-$K$ routing over $N$ experts and includes a shared expert following DeepSeekMoE~\citep{dai2024deepseekmoe}. The shared expert is applied to every token in parallel with the routed experts, and its output is added to the routed output. This captures heterogeneity-agnostic computation while allowing the routed experts to focus on source-specific variation; its ablation is reported in Appendix~\ref{appendix:shared_expert}.

\subsection{Training Objective}
HiMoE-VLA is trained with a flow-matching loss for action generation and two routing regularizers matched to the hierarchy: AS-Reg for boundary action-space specialization and HB-Reg for balanced utilization of the adjacent HB-MoE experts. The objective is
\begin{equation}
\mathcal{L} = \mathcal{L}_{\text{flow}} + \lambda_{\text{AS}} \, \mathcal{L}_{\text{AS}} + \lambda_{\text{HB}} \, \mathcal{L}_{\text{HB}},
\end{equation}
\paragraph{Flow-Matching Loss.}
We adopt the flow-matching objective~\citep{flowmatching} to model the conditional distribution of action sequences, as it provides a stable and efficient alternative to traditional diffusion training. Given an action chunk $A_t = [a_t, a_{t+1}, \ldots, a_{t+H-1}]$, flow matching defines a continuous-time trajectory that transports a noise distribution to the target action distribution:
\begin{equation}
A_t^\tau = \tau A_t + (1 - \tau)\epsilon, \quad \epsilon \sim \mathcal{N}(0, I), \quad \tau \in [0, 1].
\end{equation}
The model learns a vector field $v_\theta$ that predicts the denoising direction:
\begin{equation}
\mathcal{L}_{\text{flow}} = \mathbb{E}_{\tau,\,A_t,\,\epsilon} \left[ \big\| v_\theta(A_t^\tau, \tau, o_t, l, q_t) - \big(A_t - \epsilon \big) \big\|_2^2 \right].
\end{equation}
During training, $\tau$ is sampled from a Beta distribution following~\citep{black2024pi_0}. At inference time, future actions are generated by integrating the learned vector field from $\tau = 0$ to $\tau = 1$, starting from Gaussian noise.

\paragraph{Action-Space Regularization (AS-Reg).}
The \textbf{AS-MoE}, placed at the boundary layers, is designed to capture action-space variations. Let $c_u$ denote the action-space/embodiment identity of token $u$ (provided by dataset metadata, e.g., joint-angle vs.\ end-effector), and $\hat{r}_u \in \mathbb{R}^N$ the $\ell_2$-normalized routing probability vector of the AS-MoE router. We define the positive and anchor-excluded index sets
\begin{equation}
    P(u) = \{\, v : c_v = c_u,\ v \neq u \,\},
    \qquad
    A(u) = \{1,\dots,U\}\setminus\{u\},
\end{equation}
and adopt a supervised contrastive objective (log-inside form) over the routing distributions:
\begin{equation}
    \mathcal{L}_{\mathrm{AS}} = \frac{1}{U_+}\sum_{u=1}^{U}
        \mathds{1}\!\big[|P(u)|>0\big]\,
        \frac{-1}{|P(u)|}
        \sum_{p \in P(u)}
        \log\frac{\exp\!\big(\hat{r}_u^{\top}\hat{r}_p / \beta\big)}
                 {\sum_{v \in A(u)} \exp\!\big(\hat{r}_u^{\top}\hat{r}_v / \beta\big)},
\end{equation}
where $U_+=\sum_u\mathds{1}[|P(u)|>0]$ and $\beta{=}0.1$; if $U_+=0$, the AS-Reg term is skipped for that mini-batch. Excluding the anchor from $A(u)$ removes the dominant self-similarity term $\exp(1/\beta)$ from the denominator; the valid-anchor normalization and $1/|P(u)|$ averaging make the loss robust to batch composition and per-class frequency, so that AS-Reg aligns routing patterns within each action space and contrasts them across action spaces at the boundary layers.

\paragraph{Heterogeneity-Balancing Regularization (HB-Reg).}
At the layers adjacent to AS-MoE, the \textbf{HB-MoE} is intended to keep sparse capacity available for residual heterogeneity, such as embodiment, viewpoint, and scene variation, after action-space-specific computation has been separated. We adopt the standard load-balancing loss from DeepSeekMoE~\citep{dai2024deepseekmoe} as HB-Reg. Let $N$ be the number of experts, $K$ the top-$K$ width, $U$ the number of tokens, $s_{i,u}\!\in\![0,1]$ the router's softmax score for expert $i$ on token $u$, and $r_{i,u}\!=\!\mathds{1}\{u\!\to\!i\}$ the top-$K$ routing indicator. Define
\begin{equation}
f_i = \frac{N}{K\,U}\sum_{u=1}^{U} r_{i,u}, \quad
P_i = \frac{1}{U}\sum_{u=1}^{U} s_{i,u}, \quad
\mathcal{L}_{\mathrm{HB}} = \sum_{i=1}^{N} f_i \, P_i,
\end{equation}
so that $\mathcal{L}_{\mathrm{HB}}{=}1$ at the balanced configuration $f_i{=}1, P_i{=}1/N$; the factor $N$ keeps this reference value comparable across expert counts. Since $r_{i,u}$ is non-differentiable, $f_i$ is treated as a stop-gradient constant and gradients flow only through $s_{i,u}$, redirecting probability mass from over-used experts (large $f_i$) toward under-used ones. Statistics are computed per sequence and averaged across the batch.

In summary, AS-Reg promotes \textbf{specialization} in the AS-MoE for action-space differences, while HB-Reg promotes \textbf{balanced utilization} of the adjacent HB-MoE experts. Together with the flow-matching loss, these objectives train the hierarchy to combine source-specific routing with shared action generation.

% \section{Experiments}
% \textbf{Pre-training Dataset.}
% We pre-train HiMoE-VLA on a large-scale mixture of the Open X-Embodiment (OXE) subset~\citep{o2024open} (22.5M frames) and publicly available Aloha datasets~\citep{liu2024rdt,aloha_fine_grain,mobile_aloha} (1.6M frames), totaling 24.1M frames. 
% % This combination provides diverse embodiments, action spaces, and tasks, enabling effective cross-domain learning. 
% More details are provided in the Appendix~\ref{pretrain_dataset}

% \textbf{Implementation Details.}  
% HiMoE-VLA (4B parameters) is trained end-to-end on 16 A100 GPUs with DeepSpeed optimization. The model consumes third-person and wrist-mounted camera views, along with unified state–action vectors for both single- and dual-arm settings.  % The MoE design uses $N=32$ experts with top-$k=4$, and the auxiliary regularization coefficients are set following best practices. For the MoE gating mechanism, each token’s hidden state is fed into a linear projection to compute the expert logits. These logits are normalized via a standard softmax without temperature scaling, producing a distribution over experts. The gate selects the top-$k$ experts based on these scores, and the selected probabilities are renormalized so that their weights sum to one. This design ensures stable routing and well-scaled mixture coefficients throughout training. 
% More details are provided in Appendix~\ref{train_details}.
%===============================================================================
\section{Experiments}
We evaluate HiMoE-VLA along three axes: standard simulation performance, real-world transfer and generalization, and controlled analyses of heterogeneous co-training. The goal is to test not only final task success, but also whether the hierarchy prevents negative transfer under controlled action-space variation and shared-action observation/scene variation, while evaluating real-world performance after fine-tuning on distinct robot embodiments. The appendix provides full data, protocol, routing, and efficiency details.

\textbf{Pre-training Dataset.}
We pre-train on a large mixture of Open X-Embodiment (OXE)~\citep{o2024open} and public ALOHA datasets~\citep{liu2024rdt,aloha_fine_grain,mobile_aloha}, totaling 24.1M frames. OXE provides broad single-arm robot coverage, while ALOHA adds coordinated bimanual manipulation; Appendix~\ref{pretrain_dataset} lists the exact composition.

\textbf{Implementation Details.}
HiMoE-VLA has 4B parameters and is trained end-to-end on 16 A100 GPUs with DeepSpeed. The model consumes one third-person view, two wrist views, and unified state--action vectors covering single- and dual-arm settings. We use $N{=}32$ experts with top-$K{=}4$ routing and the AS/HB regularizers described in Sec.~3; Appendix~\ref{train_details} gives the vector layout, masks, optimization setup, and fine-tuning schedules, while Appendix~\ref{appendix:efficiency} reports overhead.

\subsection{Simulation Experiments}

\textbf{Experiment setup.}
We evaluate on CALVIN~\citep{mees2022calvin} under the $D{\rightarrow}D$ long-horizon setting, which measures chained instruction completion after fine-tuning on environment D, and on the four LIBERO~\citep{liu2023libero} suites, which test spatial, object, goal, and long-horizon generalization. In CALVIN tables, columns 1--5 report the success rate of completing at least $k$ consecutive subtasks in a five-subtask chain, and Sum. is the sum of these success rates. Appendix~\ref{benchmark} details tasks, preprocessing, baselines, and fine-tuning.

\begin{table*}[t]
    \centering
    \paperTableTight
    \caption{Simulation benchmark results. CALVIN reports the average number of consecutively completed tasks; LIBERO reports success rates (\%).}
    \begin{subtable}[t]{0.48\linewidth}
        \centering
        \begin{tabularx}{\linewidth}{@{}c*{6}{Y}@{}}
            \toprule
            \textbf{Method} & \textbf{1} & \textbf{2} & \textbf{3} & \textbf{4} & \textbf{5} & \textbf{Sum.} \\
            \midrule
            Octo      & 0.771 & 0.535 & 0.318 & 0.206 & 0.136 & 1.97 \\
            OpenVLA   & 0.716 & 0.385 & 0.180 & 0.088 & 0.042 & 1.41 \\
            RDT-1B    & 0.757 & 0.495 & 0.359 & 0.243 & 0.184 & 2.04 \\
            DeeR      & 0.853 & 0.696 & 0.549 & 0.420 & 0.312 & 2.83 \\
            MDT       & \textbf{0.937} & \underline{0.845} & \underline{0.741} & 0.644 & 0.556 & 3.72 \\
            $\pi_0$   & 0.914 & 0.830 & 0.739 & \underline{0.676} & \underline{0.599} & \underline{3.76} \\
            HiMoE-VLA & \underline{0.938} & \textbf{0.866} & \textbf{0.794} & \textbf{0.723} & \textbf{0.659} & \textbf{3.98} \\
            \midrule
            FLOWER    & 0.974 & 0.924 & 0.869 & 0.813 & 0.749 & 4.35 \\
            \textbf{+ HiMoE} & \textbf{0.979} & \textbf{0.943} & \textbf{0.904} & \textbf{0.859} & \textbf{0.801} & \textbf{4.49} \\
            \bottomrule
        \end{tabularx}
        \subcaption{CALVIN ($D\!\rightarrow\!D$).}
        \label{tab:calvin}
    \end{subtable}\hfill
    \begin{subtable}[t]{0.48\linewidth}
        \centering
        \begin{tabularx}{\linewidth}{@{}c*{5}{Y}@{}}
            \toprule
            \textbf{Method} & \textbf{Spatial} & \textbf{Object} & \textbf{Goal} & \textbf{Long} & \textbf{Avg.} \\
            \midrule
            Diffusion Policy & 78.3 & 92.5 & 68.3 & 50.5 & 72.4 \\
            Octo             & 78.9 & 85.7 & 84.6 & 51.1 & 75.1 \\
            OpenVLA          & 84.7 & 88.4 & 79.2 & 53.7 & 76.5 \\
            SpatialVLA       & 88.2 & 89.9 & 78.6 & 55.5 & 78.1 \\
            UniVLA           & 96.5 & 96.8 & 95.6 & 92.0 & 95.2 \\
            $\pi_0$          & 96.8 & \underline{98.8} & 95.8 & 85.2 & 94.2 \\
            OpenVLA-OFT      & 97.6 & 98.4 & 97.9 & \underline{94.5} & \underline{97.1} \\
            $\pi_{0.5}$      & \textbf{98.8} & 98.2 & \underline{98.0} & 92.4 & 96.8 \\
            HiMoE-VLA        & \underline{98.2} & \textbf{99.4} & \textbf{98.6} & \textbf{95.8} & \textbf{98.0} \\
            \bottomrule
        \end{tabularx}
        \subcaption{LIBERO.}
        \label{tab:libero}
    \end{subtable}
\end{table*}

\textbf{Results.}
On CALVIN, HiMoE-VLA reaches $3.98$ sum of completed subtasks, outperforming the strongest prior generalist $\pi_0$ ($3.76$) and all other non-FLOWER baselines. Replacing FLOWER's dense action expert with HiMoE further improves $4.35$ to $\mathbf{4.49}$, indicating that the architecture is compatible with a different training recipe. On LIBERO, HiMoE-VLA obtains the best average score (98.0\%) and leads on Object, Goal, and Long, while $\pi_{0.5}$ remains slightly stronger on Spatial.
% Compared to OpenVLA-OFT, the previous state-of-the-art (97.1\%), HiMoE-VLA delivers consistent gains across all four suites: +0.6\% on Spatial, +1.0\% on Object, +0.7\% on Goal, and +0.3\% on Long. 
These results establish the base policy performance before the controlled heterogeneity studies in Sec.~\ref{sec:model_analysis}.

\subsection{Real-World Experiments}
We evaluate on xArm7 single-arm and ALOHA dual-arm robots.

\textbf{Experiment setup.}
We test \textit{Fruit-to-Plate}, \textit{Cup-in-Cup}, and \textit{Block-on-Block} on xArm7, and \textit{Fold-Shorts}, \textit{Cup-Handover}, and \textit{Scoop} on ALOHA. These tasks cover pick-place, insertion, stacking, bimanual handover, scooping, and deformable-object folding. We report stage-level success and generalization to unseen distractors and novel objects/garments; Appendix~\ref{benchmark} gives demonstrations, settings, and trial counts.

\begin{table*}[t]
    \centering
    \paperTableTight
    \caption{Real-world evaluation on xArm7 (single-arm) and ALOHA (dual-arm) robots. (a) xArm7 per-stage success rates; (b) ALOHA per-stage success rates.}
    \label{tab:realworld}
    \begin{subtable}[t]{0.48\linewidth}
        \centering
        \begin{tabularx}{\linewidth}{@{}Y*{7}{c}@{}}
            \toprule
            \multirow{2}{*}{Method}
                & \multicolumn{2}{c}{Fruit-to-Plate}
                & \multicolumn{2}{c}{Cup-in-Cup}
                & \multicolumn{2}{c}{Block-on-Block}
                & Avg. \\
            \cmidrule(lr){2-3} \cmidrule(lr){4-5} \cmidrule(lr){6-7}
                & Pick & Place & Pick & Insert & Pick & Stack & \\
            \midrule
            Octo-Base & 31.3 & 18.8 & 33.3 & 16.7 & 16.7 & 0.0 & 19.3 \\
            OpenVLA   & 37.5 & 25.0 & 27.8 & 16.7 & 22.2 & 0.0 & 21.2 \\
            CogACT    & 65.6 & 59.4 & \underline{77.8} & \underline{63.9} & 69.4 & \underline{33.3} & 61.5 \\
            $\pi_0$   & \underline{68.8} & \underline{62.5} & \underline{77.8} & 61.1 & \underline{72.2} & \underline{33.3} & \underline{62.5} \\
            HiMoE-VLA & \bf81.3 & \bf75.0 & \bf88.9 & \bf72.2 & \bf83.3 & \bf50.0 & \bf75.0 \\
            \bottomrule
        \end{tabularx}
        \subcaption{xArm7 single-arm tasks.}
        \label{tab:xarm}
    \end{subtable}\hfill
    \begin{subtable}[t]{0.50\linewidth}
        \centering
        \begin{tabularx}{\linewidth}{@{}Y*{8}{c}@{}}
            \toprule
            \multirow{2}{*}{Method}
                & \multicolumn{2}{c}{Cup-Handover}
                & \multicolumn{3}{c}{Scoop}
                & \multicolumn{2}{c}{Fold-Shorts}
                & Avg. \\
            \cmidrule(lr){2-3} \cmidrule(lr){4-6} \cmidrule(lr){7-8}
                & Grasp & Trans & Place & Scoop & Pour & Once & Twice & \\
            \midrule
            ACT       & 40.0 & 0.0 & 73.3 & 6.6 & 0.0 & 20.0 & 6.6 & 20.9 \\
            RDT-1B    & 66.6 & \underline{13.3} & \underline{93.3} & 40.0 & 20.0 & 53.3 & 46.6 & 47.5 \\
            $\pi_0$   & \bf80.0 & \underline{13.3} & \underline{93.3} & \underline{46.6} & \underline{26.6} & \underline{66.6} & \underline{53.3} & \underline{54.2} \\
            HiMoE-VLA & \bf80.0 & \bf26.6 & \bf100.0 & \bf53.3 & \bf40.0 & \bf80.0 & \bf66.6 & \bf63.7 \\
            \bottomrule
        \end{tabularx}
        \subcaption{ALOHA dual-arm tasks (Trans=Transfer).}
        \label{tab:aloha}
    \end{subtable}
\end{table*}

\begin{table}[t]
    \centering
    \paperTable
    \caption{Real-world generalization evaluation on single-arm (xArm7) and dual-arm (ALOHA) tasks under two scenarios: \textit{Distractor Objects} (unseen distractors) and \textit{Novel Objects} (previously unseen items). Entries are success rates (\%) averaged over physical trials; Appendix~\ref{benchmark} reports the trial counts.}
    \label{tab:gene}
    \begin{tabularx}{0.7\linewidth}{@{}c*{6}{Y}@{}}
        \toprule
        \multirow{2}{*}{Method}
            & \multicolumn{3}{c}{Single-Arm}
            & \multicolumn{3}{c}{Dual-Arm} \\
        \cmidrule(lr){2-4} \cmidrule(lr){5-7}
            & Distractor & Novel & Avg.
            & Distractor & Novel & Avg. \\
        \midrule
        OpenVLA        &19.4 &15.6 &17.6 &- &- &- \\
        CogACT         &52.8 &50.0 &51.5 &- &- &- \\
        RDT-1B         &- &- &- &28.9 &\underline{26.7} &27.8 \\
        $\pi_0$        &\underline{58.3} &\underline{53.1} &\underline{55.9} &\underline{40.0} &\underline{26.7} &\underline{33.4} \\
        HiMoE-VLA      &\bf69.4 &\bf65.6 &\bf67.6 &\bf53.3 &\bf46.7 &\bf50.0 \\
        \bottomrule
    \end{tabularx}
\end{table}

\begin{figure*}[t]
    \centering
    \includegraphics[width=0.99\linewidth]{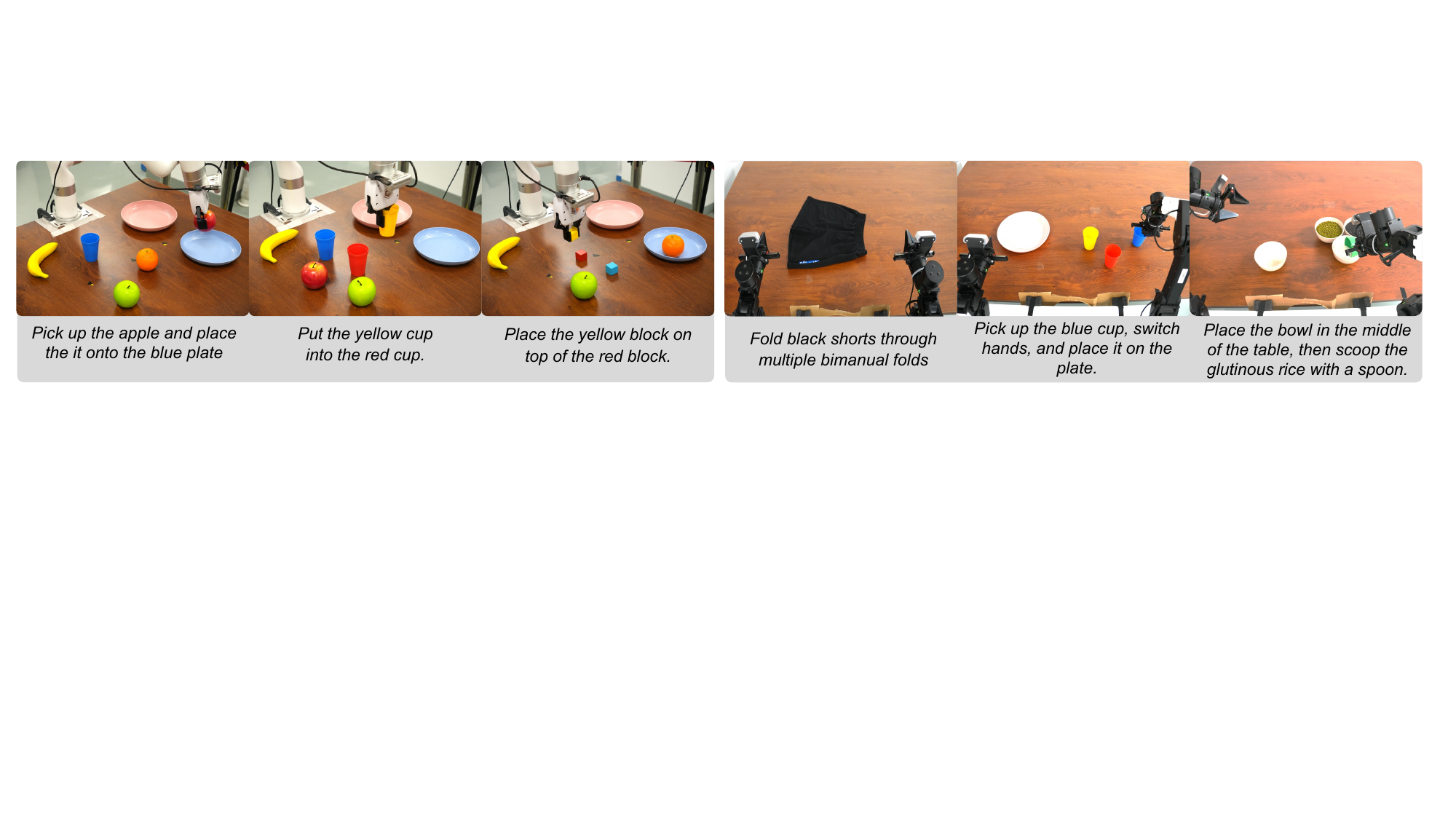}
    \caption{Qualitative examples of real-world executions on (top row) the single-arm xArm7 and (bottom row) the dual-arm ALOHA robot. The snapshots cover representative stages across tasks such as \textit{Fruit-to-Plate}, \textit{Block-on-Block}, \textit{Cup-Handover}, and \textit{Scoop}.}
    \label{fig:xarm_aloha_examples}
\end{figure*}

\textbf{Results.}
Table~\ref{tab:realworld} shows that HiMoE-VLA achieves the best average success on xArm7 (75.0\%) and ALOHA (63.7\%), improving over $\pi_0$ by 12.5 and 9.5 points. The largest gains appear in coordinated transfer, pouring, and folding stages, where embodiment-specific control and long-horizon coordination are likely to matter. Table~\ref{tab:gene} further reports higher success rates than $\pi_0$ on both single-arm (67.6\% vs.\ 55.9\%) and dual-arm settings (50.0\% vs.\ 33.4\%). Representative rollouts appear in Fig.~\ref{fig:xarm_aloha_examples} and Appendix Figs.~\ref{fig:xarm2}--\ref{fig:aloha}.

\begin{table}[t]
    \vspace{-2pt}
    \centering
    \paperTable
    \caption{Main CALVIN ablations.}
    \label{tab:ablation-all}
    \begin{tabularx}{0.72\linewidth}{@{}cc*{6}{Y}@{}}
        \toprule
        \textbf{Study} & \textbf{Setting} & \textbf{1} & \textbf{2} & \textbf{3} & \textbf{4} & \textbf{5} & \textbf{Sum.} \\
        \midrule
        \multirow{3}{*}{Opt.} & w/o warm-up  & 0.917 & 0.832 & 0.753 & 0.698 & 0.627 & 3.827 \\
                              & w/o pretrain & 0.928 & 0.845 & 0.752 & 0.686 & 0.615 & 3.826 \\
                              & Full         & \bf0.938 & \bf0.866 & \bf0.794 & \bf0.723 & \bf0.659 & \bf3.980 \\
        \midrule
        \multirow{3}{*}{Hetero.} & Sep.\ Heads & 0.914 & 0.833 & 0.753 & 0.696 & 0.631 & 3.827 \\
                                 & GR00T-Like  & 0.913 & 0.835 & 0.764 & 0.702 & 0.642 & 3.856 \\
                                 & HiMoE       & \bf0.943 & \bf0.864 & \bf0.797 & \bf0.734 & \bf0.674 & \bf4.012 \\
        \bottomrule
    \end{tabularx}
\end{table}

\subsection{Model Analysis and Ablations}
\label{sec:model_analysis}
We ablate optimization, action-space heterogeneity, hierarchical design, and shared-action observation/scene heterogeneity. For CALVIN action-space experiments, D denotes environment D with joint-angle actions, while ABC denotes environments A/B/C with end-effector (EEF) actions; ABC+D mixes these two action parameterizations. Scalar entries are CALVIN Sum. scores unless otherwise stated. The main text includes the controls needed to support the core claims; Appendix~\ref{appendix:ablation_protocol} defines all variants and reports full per-step results.
\begin{wraptable}{r}{0.42\linewidth}
    \vspace{-2pt}
    \centering
    \paperTableTight
    \caption{Action-space co-training on CALVIN.}
    \label{tab:hetero-cotraining}
    \begin{tabularx}{\linewidth}{@{}cYY@{}}
        \toprule
        \textbf{Method} & \textbf{D} & \textbf{ABC+D} \\
        \midrule
        $\pi_0$       & 3.806 & 3.547 (\textbf{-0.259}) \\
        Ours w/o MoE  & 3.819 & 3.777 (\textbf{-0.042}) \\
        Full (HiMoE)  & 3.826 & 4.012 (\textbf{+0.186}) \\
        \bottomrule
    \end{tabularx}
    \vspace{-2pt}
\end{wraptable}
\textbf{Effect of Warm-up and Pretraining.}
The optimization rows in Table~\ref{tab:ablation-all} show that removing MoE warm-up or pretraining degrades CALVIN-D fine-tuning, confirming the value of pretrained representations and short expert adaptation. Appendix~\ref{warmup} details the two-stage warm-up that first adapts MoE parameters before full fine-tuning.

\begin{wraptable}{r}{0.3\linewidth}
    \vspace{-2pt}
    \centering
    \paperTableTight
    \caption{HiMoE component ablation on CALVIN ABC+D co-training.}
    \label{tab:hb-moe}
    \begin{tabularx}{\linewidth}{@{}YY@{}}
        \toprule
        \textbf{Setting} & \textbf{Sum.} \\
        \midrule
        w/o MoE        & 3.777 \\
        Full-HB-MoE    & \underline{3.901} \\
        w/o AS-MoE     & 3.873 \\
        w/o HB-MoE     & 3.836 \\
        w/o Reg        & 3.835 \\
        Single-MoE+Reg & 3.813 \\
        Full           & \bf4.012 \\
        \bottomrule
    \end{tabularx}
    \vspace{-2pt}
\end{wraptable}

\textbf{Comparison with Other Methods on Handling Heterogeneous Action Spaces.}
The heterogeneity rows compare HiMoE with separate heads and a GR00T-style embodiment indicator under CALVIN-ABC-EEF + CALVIN-D-Joint co-training. HiMoE performs best without manual embodiment heads, suggesting that adaptive expert selection better balances specialization and sharing. Appendix~\ref{appendix:training_pipeline} audits the data/loss masking pipeline used to avoid padding artifacts.

\textbf{Evaluation on Ability of HiMoE to Handle Heterogeneous Data.}
Table~\ref{tab:hetero-cotraining} compares from-scratch training on CALVIN-D joint-angle data only with CALVIN-ABC (EEF) + CALVIN-D (joint-angle) co-training. Heterogeneous co-training degrades $\pi_0$ and the dense variant, whereas full HiMoE turns the mixture into a gain. Cross-action fine-tuning in Appendix~\ref{appendix:cross-action-ft} shows the same trend, improving EEF and joint evaluation by $+0.105$ and $+0.238$ over isolated fine-tuning.

\textbf{Role of Hierarchical MoE Components.}
Table~\ref{tab:hb-moe} shows that removing MoE layers, AS-MoE, HB-MoE, or regularization consistently hurts performance, and a single non-hierarchical MoE remains weaker; full per-step results are provided in Table~\ref{tab:hb-moe-full}. The gap between \textit{Full-HB-MoE} (3.901) and full HiMoE (4.012) supports placing AS-MoE at boundary layers rather than merely adding sparse capacity.

Appendix controls further rule out confounds: removing the shared expert drops CALVIN ABC+D co-training from $4.012$ to $3.955$ (Table~\ref{tab:ablation_calvin}); a parameter-matched dense model reaches only $3.801$ despite more active parameters (Table~\ref{tab:param_scaling}); and removing loss/data masks in heterogeneous co-training degrades the dense baseline from $3.777$ to $3.696/3.626$ (Table~\ref{tab:ablation_masks}).

\begin{wraptable}{r}{0.42\linewidth}
    \vspace{-5pt}
    \centering
    \paperTableTight
    \caption{Sensor/scene co-training with a shared EEF action space.}
    \label{tab:sensor-hetero}
    \begin{tabularx}{\linewidth}{@{}cYY@{}}
        \toprule
        \textbf{Method} & \textbf{CALVIN} & \textbf{+LIBERO} \\
        \midrule
        $\pi_0$        & 3.776 & 3.504 (\textbf{-0.272}) \\
        w/o MoE        & 3.788 & 3.665 (\textbf{-0.123}) \\
        Standard MoE   & 3.808 & 3.862 (\textbf{+0.054}) \\
        Full (HiMoE)   & 3.819 & \bf 3.966 (\textbf{+0.147}) \\
        \bottomrule
    \end{tabularx}
    \vspace{-5pt}
\end{wraptable}
\textbf{Evaluating Sensor and Scene Heterogeneity Under a Shared Action Space.}
To reduce action-space confounding, we co-train CALVIN-D and LIBERO under a shared EEF action parameterization. Because the two benchmarks still differ in task distribution, visual domain, and scene layout, this experiment should be viewed as a shared-action test of non-action heterogeneity rather than a pure sensor-only intervention. Table~\ref{tab:sensor-hetero} shows negative transfer for $\pi_0$ and dense models, a small gain for standard MoE, and a larger gain for full HiMoE ($+0.147$), supporting the hierarchy beyond action-space differences.

\textbf{Scaling the Number of Experts.}
Appendix Table~\ref{tab:experts-topk} shows that $N{=}32,K{=}4$ performs best; larger $N$ gives diminishing returns and $K{=}8$ is unstable. The appendix also reports modest efficiency overhead (about $7\%$ training cost and $0.195$s/action inference for $N{=}32,K{=}4$) and routing heatmaps consistent with AS-MoE action-space specialization and HB-MoE balanced utilization.

%===============================================================================
\section{Conclusion}
We presented HiMoE-VLA, a VLA framework that uses a hierarchical MoE action module to separate action-space-specific computation and allocate balanced capacity to residual sources of robot-data variation, including observations, scenes, and embodiments. Across CALVIN, LIBERO, and real xArm7/ALOHA tasks, HiMoE-VLA improves over strong baselines; controlled co-training studies further show that the hierarchy mitigates negative transfer under action-space and shared-action observation/scene variation beyond a pure capacity effect.

\textbf{Limitations:} HiMoE-VLA assumes that each sample can be mapped into a unified state--action interface; failures may arise when annotations are missing, or mobile-manipulation settings differ substantially from our evaluated robots. Our experiments cover CALVIN, LIBERO, xArm7, and ALOHA but not larger multi-robot mixtures, and we do not yet quantify safety or calibration under long-horizon distribution shift. Future work should further reduce routing and cross-attention overhead and broaden evaluation.

%===============================================================================
\clearpage
% % The acknowledgments are automatically included only in the final and preprint versions of the paper.
% \acknowledgments{If a paper is accepted, the final camera-ready version will (and probably should) include acknowledgments. All acknowledgments go at the end of the paper, including thanks to reviewers who gave useful comments, to colleagues who contributed to the ideas, and to funding agencies and corporate sponsors that provided financial support.}

%===============================================================================

\bibliography{example}  % .bib

%===============================================================================
% Appendix stubs. These sections are referenced from the main text
% (Appendix~\ref{appendix:transfer}, \ref{train_details}, etc.).
% Replace the placeholder bodies with the actual appendix content as needed.
\clearpage
\appendix

\section{Dataset and Evaluation}
\subsection{Pretraining Dataset}
\label{pretrain_dataset}
Our pre-training dataset is constructed by combining subsets of Open X-Embodiment (OXE) and publicly available ALOHA datasets, yielding a total of 24.1M frames. The detailed data mixture is listed in Table~\ref{tab:data_mix}.

\textbf{OXE dataset.} OXE~\citep{o2024open} aggregates over 1 million real-world trajectories collected from 60 datasets across 22 distinct robot embodiments. Following prior works such as Octo~\citep{team2024octo} and OpenVLA~\citep{kim2024openvla}, we adopt a subset containing 22.5M frames, chosen to balance scale and diversity while ensuring compatibility with our training pipeline. This subset spans a wide range of single-arm robots and tasks, providing strong coverage of heterogeneous embodiments and action spaces.

\textbf{ALOHA datasets.} To complement OXE, we incorporate demonstrations from three high-quality, publicly available ALOHA datasets~\citep{liu2024rdt,aloha_fine_grain,mobile_aloha}, contributing 1.6M frames in total. Compared to OXE, they emphasize coordinated bimanual actions and higher-fidelity manipulation skills, substantially enriching the diversity of our training corpus.

\begin{table}[ht]
    \centering
    \paperTable
    \caption{Our training data mixture using datasets from the Open X-Embodiment dataset~\citep{o2024open} and ALOHA dataset~\citep{liu2024rdt,aloha_fine_grain,mobile_aloha}.}
    \begin{tabularx}{0.5\linewidth}{@{}Yc@{}}
        \toprule
        \textbf{Dataset} & \textbf{Ratio} \\
        \midrule
        Fractal~\citep{brohan2022rt} & 23.9\% \\
        Kuka~\citep{kalashnikov2018qt} & 12.9\% \\
        Bridge~\citep{ebert2021bridge, walke2023bridgedata} & 11.9\% \\
        Taco Play~\citep{rosete2022tacorl, mees23hulc2} & 2.7\% \\
        Jaco Play~\citep{dass2023jacoplay} & 0.4\% \\
        Berkeley Cable Routing~\citep{luo2023multistage} & 0.2\% \\
        Roboturk~\citep{mandlekar2019scaling} & 2.1\% \\
        Viola~\citep{zhu2022viola} & 0.9\% \\
        Berkeley Autolab UR5~\citep{BerkeleyUR5Website} & 1.1\% \\
        Toto~\citep{zhou2023train} & 1.9\% \\
        Stanford Hydra Dataset~\citep{belkhale2023hydra}  & 4.5\% \\
        Austin Buds Dataset~\citep{zhu2022bottom}  & 0.2\% \\
        NYU Franka Play Dataset~\citep{cui2022play}  & 0.7\% \\
        Furniture Bench Dataset~\citep{heo2023furniturebench}  & 2.5\% \\
        UCSD Kitchen Dataset~\citep{ucsd_kitchens}  & \textless 0.1\% \\
        Austin Sailor Dataset~\citep{nasiriany2022sailor}  & 2.2\% \\
        Austin Sirius Dataset~\citep{liu2022robot}  & 1.8\% \\
        DLR EDAN Shared Control~\citep{quere_shared_2020}  & \textless 0.1\% \\
        IAMLab CMU Pickup Insert~\citep{saxena2023multiresolution}  & 0.9\% \\
        UTAustin Mutex~\citep{shah2023mutex} & 2.3\% \\
        Berkeley Fanuc Manipulation~\citep{fanuc_manipulation2023} & 0.8\% \\
        CMU Stretch~\citep{mendonca2023structured} & 0.2\% \\
        BC-Z~\citep{jang2022bc} & 6.9\% \\
        FMB Dataset~\citep{luo2024fmb}  & 7.2\% \\
        DobbE~\citep{shafiullah2023dobbe}  & 1.4\% \\
        ALOHA Dataset~\citep{liu2024rdt,aloha_fine_grain,mobile_aloha} & 10.4\% \\
        \bottomrule
        \end{tabularx}%
        \label{tab:data_mix}
\end{table}

\subsection{Evaluation Benchmarks }
\label{benchmark}
\paragraph{CALVIN benchmark.}
CALVIN~\citep{mees2022calvin} is a benchmark for evaluating instruction-conditioned policies in long-horizon tabletop manipulation tasks using a Franka Panda arm. It comprises 34 tasks spanning from simple pick-and-place to articulated object manipulation and provides four simulated environments, denoted A, B, C, and D. In the standard $D \!\rightarrow\! D$ setting, policies are trained on a limited subset of demonstrations from environment D and evaluated on held-out instruction sequences in the same environment. In our heterogeneous co-training experiments, ABC denotes demonstrations from environments A/B/C using the end-effector (EEF) action parameterization, while D denotes demonstrations from environment D using joint-angle actions unless otherwise specified.

CALVIN evaluates long-horizon control with five-subtask instruction chains. Column $k\in\{1,\ldots,5\}$ reports the fraction of evaluation chains in which the policy completes at least the first $k$ consecutive subtasks. The scalar Sum. metric is the sum of these five success rates, equivalently the average number of completed subtasks per five-subtask chain; this is the standard primary CALVIN score used throughout our tables. This setup evaluates the model’s ability to generalize to novel instruction compositions under restricted data conditions. For fair comparison, we include Octo~\citep{team2024octo}, OpenVLA~\citep{kim2024openvla}, RDT-1B~\citep{liu2024rdt}, DeeR~\citep{DeeR}, MDT~\citep{MDT}, and $\pi_0$~\citep{black2024pi_0} as baselines. For DeeR and MDT, we directly report the results from their original papers. For Octo, OpenVLA, RDT-1B, and $\pi_0$, we adopt their released pre-trained weights and perform fine-tuning on CALVIN-D following their official training procedures to ensure fairness and reproducibility.

\paragraph{LIBERO benchmark.}
LIBERO~\citep{liu2023libero} is a simulation suite designed to evaluate lifelong learning and generalization in robotic manipulation. It contains four task suites—LIBERO-Spatial, LIBERO-Object, LIBERO-Goal, and LIBERO-Long—each comprising 10 tasks with 50 human-teleoperated demonstrations per task. These suites test complementary aspects of generalization: spatial reasoning (Spatial), object-level transfer (Object), goal-directed adaptability (Goal), and long-horizon planning (Long). Following prior works such as OpenVLA~\citep{kim2024openvla}, we preprocess demonstrations by removing failure cases, standardizing image inputs, and ensuring consistent trajectory formatting. In our experiments, we perform supervised fine-tuning within each task suite using the successful demonstrations and evaluate policies on held-out task episodes. We compare against strong baselines, including Diffusion Policy~\citep{dp}, Octo~\citep{team2024octo}, OpenVLA~\citep{kim2024openvla}, SpatialVLA~\citep{spatialvla}, OpenVLA-OFT~\citep{openvla-oft}, UniVLA~\citep{univla}, and $\pi_0$~\citep{black2024pi_0}, where reported results are either taken directly from their papers or reproduced under their released implementations.

\paragraph{Real-world xArm7 benchmark.}
We conduct real-world evaluations on an xArm7 robot (7-DoF manipulator with a 1-DoF gripper) across three tasks:
(1) \textit{Fruit-to-Plate} — placing fruits (apple, orange) onto colored plates (blue, pink), e.g., “Pick up the apple and place it onto the blue plate”;
(2) \textit{Cup-in-Cup} — inserting one colored cup (red, yellow, blue) into another, e.g., “Put the yellow cup into the red cup”;
(3) \textit{Block-on-Block} — stacking one colored block onto a differently colored block, e.g., “Place the yellow block on top of the red block.”
Each task is decomposed into sub-stages (e.g., Pick/Place, Pick/Insert, Pick/Stack) for fine-grained evaluation.
We collect 320 teleoperated demonstrations in total: 80 (\textit{Fruit-to-Plate}), 120 (\textit{Cup-in-Cup}), and 120 (\textit{Block-on-Block}), with 20 demonstrations per configuration.

\textit{In-distribution evaluation.}
\textbf{Fruit-to-Plate}: 4 settings $\times$ 4 trials/setting = 16 trials in total, where each “setting” is a fruit–plate pairing from \{apple, orange\} $\times$ \{blue, pink\}.
\textbf{Cup-in-Cup}: 6 settings $\times$ 3 trials/setting = 18 trials, where each “setting” is an \emph{ordered} inner$\rightarrow$outer color pair from \{red, yellow, blue\} with distinct colors (i.e., $3 \times 2 = 6$ ordered pairs).
\textbf{Block-on-Block}: 6 settings $\times$ 3 trials/setting = 18 trials, where each “setting” is an \emph{ordered} top$\rightarrow$bottom color pair (distinct) from \{red, yellow, blue\}.

\textit{Generalization tests.}
(1) \textbf{Distractors in \textit{Cup-in-Cup}}: 6 settings (the same 6 inner$\rightarrow$outer color pairs as above) $\times$ 3 trials/setting = 18 trials, with an unseen distractor (e.g., a pomegranate or a green cup) placed in the scene.  
(2) \textbf{Novel objects in \textit{Fruit-to-Plate}}: 4 settings $\times$ 4 trials/setting = 16 trials.  
Here, “4$\times$4” means that we test four novel configurations — placing a pomegranate onto a blue plate, a pomegranate onto a pink plate, an apple onto a purple plate, and an orange onto a purple plate — with each configuration repeated for 4 trials.

\paragraph{Real-world ALOHA benchmark.}
We further evaluate on the ALOHA robot (dual-arm, 14-DoF) with three tasks:
(1) \textit{Fold-Shorts} — folding a pair of shorts (50 teleoperated demonstrations), e.g., “Fold black shorts through multiple bimanual folds”;
(2) \textit{Cup-Handover} — the right arm grasps a colored cup (red, yellow, blue) and hands it to the left arm to place on a plate (60 demos per color; 180 total), e.g., “Pick up the blue cup, switch hands, and place it on the plate”;
(3) \textit{Scoop} — the left arm places a bowl centrally, then the right arm uses a spoon to scoop materials (mung beans, black rice, sticky rice) into the bowl (40 demos per material; 120 total), e.g., “Place the bowl in the middle of the table, then scoop the glutinous rice with a spoon.”
Altogether, 350 demonstrations are collected.

\textit{In-distribution evaluation.}
\textbf{Fold-Shorts}: 1 setting $\times$ 15 trials = 15 trials.
\textbf{Cup-Handover}: 3 settings (one per cup color) $\times$ 5 trials/setting = 15 trials.
\textbf{Scoop}: 3 settings (one per material type) $\times$ 5 trials/setting = 15 trials.

\textit{Generalization tests.}
(1) \textbf{Distractors in \textit{Scoop}}: 3 settings (the same three material types) $\times$ 3 trials/setting = 9 trials, with unseen distractors (e.g., banana or green apple) added to the scene.
(2) \textbf{Novel garment in \textit{Fold-Shorts}}: 1 setting (previously unseen shorts) $\times$ 15 trials = 15 trials.

\section{Implementation Details}
\label{train_details}

\paragraph{Model scale and training setup.}
Our proposed HiMoE-VLA model contains approximately 4B parameters and is trained end-to-end on 16 NVIDIA A100 GPUs (40GB each) for 100k steps with a global batch size of 256. Training takes around 4 days with DeepSpeed optimization, and we adopt the LeRobot data-loading framework to ensure efficient and scalable handling of large heterogeneous datasets.

\paragraph{Input modalities.}
The visual encoder consumes one third-person camera view together with two wrist-mounted views. When a view is unavailable in a dataset, the corresponding channel is zero-padded and masked using attention masks, ensuring a consistent input format. For state and action inputs, we construct a unified vector representation that jointly accommodates both joint-angle and end-effector signals. In single-arm demonstrations, the available arm is mapped to the right-arm channel, while the left-arm channel is zero-padded with masks to preserve compatibility with dual-arm settings. 
All heterogeneous actions and states are mapped into a fixed 24-dimensional vector, consisting of 8-dimensional end-effector actions and 16-dimensional joint angles. For states, a validity mask is concatenated to indicate which segments are active. For actions, unavailable dimensions are zero-padded, and the loss mask described in Appendix~\ref{appendix:training_pipeline} restricts supervision to valid action dimensions.

\paragraph{Mixture-of-Experts design.}
We set the number of experts to $N=32$ with top-$K$ routing of $K=4$. As shown in Table~\ref{tab:experts-topk}, this configuration consistently outperforms alternative settings in terms of both average performance and stability, striking a favorable balance between model capacity and computational efficiency. To encourage effective expert utilization and hierarchical abstraction, we introduce two auxiliary regularizations: an Action-Space regularization term with coefficient $\lambda_{\text{AS}}=0.002$, and a Heterogeneity-Balancing regularization term with coefficient $\lambda_{\text{HB}}=0.001$. These choices follow best practices for balancing specialization and generalization in MoE architectures.

\begin{table}[ht]
    \centering
    \paperTable
    \caption{Ablation on the number of experts $N$ and top-$K$ routing on CALVIN. $N{=}32$, $K{=}4$ gives the best trade-off between specialization and stability.}
    \label{tab:experts-topk}
    \begin{tabular}{@{}cc*{6}{c}@{}}
        \toprule
        $\bm{K}$ & $\bm{N}$ & \textbf{1} & \textbf{2} & \textbf{3} & \textbf{4} & \textbf{5} & \textbf{Sum.} \\
        \midrule
        \multirow{4}{*}{2}
            & 2  & 0.895 & 0.811 & 0.757 & 0.712 & 0.648 & 3.823 \\
            & 4  & 0.901 & 0.814 & 0.761 & 0.715 & 0.653 & 3.844 \\
            & 8  & 0.910 & 0.827 & 0.768 & 0.722 & 0.669 & 3.896 \\
            & 16 & 0.920 & 0.846 & 0.781 & 0.733 & 0.671 & 3.951 \\
        \midrule
        \multirow{4}{*}{4}
            & 8  & 0.921 & 0.847 & 0.776 & 0.715 & 0.657 & 3.916 \\
            & 16 & \underline{0.923} & 0.846 & 0.774 & 0.729 & \bf0.682 & 3.954 \\
            & 32 & \bf0.943 & \bf0.864 & \bf0.797 & \underline{0.734} & \underline{0.674} & \bf4.012 \\
            & 64 & 0.919 & \underline{0.854} & \underline{0.785} & \bf0.738 & 0.672 & \underline{3.968} \\
        \midrule
        \multirow{2}{*}{8}
            & 16 & 0.911 & 0.773 & 0.637 & 0.546 & 0.458 & 3.325 \\
            & 32 & 0.897 & 0.794 & 0.719 & 0.673 & 0.612 & 3.695 \\
        \bottomrule
    \end{tabular}
\end{table}

\paragraph{Optimization and fine-tuning.}
We adopt the AdamW optimizer with an initial learning rate of $2.5\times10^{-5}$, weight decay of $1\times10^{-4}$, and a cosine decay schedule. The learning rate is linearly warmed up for the first 1k steps, followed by exponential decay until 30k steps with a final floor of $2.5\times10^{-6}$. 
For fine-tuning, we adapt the batch size and number of steps to each benchmark. On \textsc{CALVIN}, we use a global batch size of 32 for 40k steps. On \textsc{LIBERO}, we fine-tune each suite separately: \textsc{Long}, \textsc{Goal}, and \textsc{Object} use batch size 64 for 40k, 45k, and 45k steps, respectively, while \textsc{Spatial} uses batch size 32 for 35k steps. For real-world experiments, we fine-tune on both \textsc{xArm7} and \textsc{ALOHA} robots with batch size 64 for 50k steps.

\paragraph{Cross-layer KV integration.}
At each transformer layer $l$, HiMoE receives the key–value pairs 
$\{K^{\mathrm{V}}_l, V^{\mathrm{V}}_l\}$ from the corresponding VLM layer, which are concatenated with the locally computed 
$\{K^{\mathrm{H}}_l, V^{\mathrm{H}}_l\}$ of the action expert:
\[
\tilde{K}_l = \big[ K^{\mathrm{H}}_l ; K^{\mathrm{V}}_l \big], 
\qquad 
\tilde{V}_l = \big[ V^{\mathrm{H}}_l ; V^{\mathrm{V}}_l \big].
\]
The query $Q^{\mathrm{H}}_l$ then attends to the fused representation:
\[
\text{Attn}_l(Q^{\mathrm{H}}_l, \tilde{K}_l, \tilde{V}_l) 
= \text{softmax}\!\left(\frac{Q^{\mathrm{H}}_l \tilde{K}_l^\top}{\sqrt{d_k}}\right)\tilde{V}_l.
\]

This design enables each HiMoE layer to directly condition on semantically aligned signals from its VLM counterpart, instead of relying solely on the final-layer representation. During inference, we further employ a KV cache to reuse the VLM’s intermediate keys and values, substantially accelerating policy rollout without degrading performance.

\section{More Analysis}
\subsection{Challenges in Transferability Across Heterogeneous Action Spaces}
\label{appendix:transfer}
Data from different action spaces are largely non-transferable due to fundamental differences in physical interpretation and kinematic structures.

\textbf{Distinct Physical Meanings.}
End-effector (EEF) actions and joint-angle actions represent distinct physical quantities. EEF actions typically define the gripper's Cartesian pose (position and orientation), whereas joint-angle actions specify individual joint rotations. Consequently, they operate in disjoint coordinate systems and adhere to different physical constraints.

\textbf{Kinematic Incompatibility.}
Even within the same action type, robots often possess varying Degrees of Freedom (DoFs) and kinematic chains. For instance, a 6-DoF arm and a 7-DoF arm executing the same end-effector motion will exhibit divergent joint-angle trajectories due to their structural differences. As a result, joint-space data cannot be directly aligned across different embodiments.

Driven by these distinct meanings and embodiment-dependent kinematics, EEF and joint-angle actions reside in incompatible domains with divergent data distributions. This significant distributional mismatch fundamentally limits the direct transferability of policies learned in one action space to another.

\subsection{Effectiveness of the Shared Expert Mechanism}
\label{appendix:shared_expert}

The shared expert captures common knowledge that is agnostic to data heterogeneity. This mechanism allows other experts to focus solely on input-specific variations (e.g., sensor configurations) rather than repeatedly learning redundant patterns. Consequently, the shared expert facilitates more efficient knowledge sharing and stabilizes the specialization of the remaining experts. This design aligns with recent advancements in MoE architectures, such as DeepSeekMoE~\citep{dai2024deepseekmoe} and LlamaMoE~\citep{zhu2024llamamoe}.

To verify its contribution, we conducted an ablation study by removing the shared expert from the architecture. As shown in Table~\ref{tab:ablation_calvin}, the performance dropped moderately, indicating that this component is essential for achieving stable specialization and stronger overall performance.

\begin{table}[ht]
    \centering
    \paperTable
    \caption{Shared-expert ablation under CALVIN-ABC (EEF) + CALVIN-D (joint-angle) co-training.}
    \label{tab:ablation_calvin}
    \begin{tabular}{@{}c*{6}{c}@{}}
        \toprule
        \textbf{Method} & \textbf{1} & \textbf{2} & \textbf{3} & \textbf{4} & \textbf{5} & \textbf{Sum.} \\
        \midrule
        w/o shared expert & 0.921 & 0.846 & 0.779 & 0.732 & \textbf{0.677} & 3.955 \\
        Full & \textbf{0.943} & \textbf{0.864} & \textbf{0.797} & \textbf{0.734} & 0.674 & \textbf{4.012} \\
        \bottomrule
    \end{tabular}
\end{table}

\subsection{Ablation Protocol and Variant Definitions}
\label{appendix:ablation_protocol}

Our ablations span the two regimes in which VLA models are commonly deployed. First, we fine-tune the pre-trained HiMoE-VLA on CALVIN-D (optimization rows of Table~\ref{tab:ablation-all}) and additionally study cross-action-space fine-tuning on CALVIN-ABC (EEF) plus CALVIN-D (joint-angle) in Appendix~\ref{appendix:cross-action-ft}. Second, we co-train from scratch on heterogeneous sources: CALVIN-ABC plus CALVIN-D for action-space heterogeneity (heterogeneity rows of Table~\ref{tab:ablation-all} and Tables~\ref{tab:hetero-cotraining},~\ref{tab:hb-moe}), and CALVIN environment D plus LIBERO under a shared EEF action parameterization for sensor/scene heterogeneity (Table~\ref{tab:sensor-hetero}). This protocol ensures that the reported gains are not tied to a single robot, benchmark, or training regime.

For the component ablation in Table~\ref{tab:hb-moe}, \textit{w/o MoE} replaces all MoE layers with dense Transformer blocks; \textit{Full-HB-MoE} replaces every MoE slot with an HB-MoE layer; \textit{w/o AS-MoE} removes only the boundary AS-MoE layers; \textit{w/o HB-MoE} removes only the adjacent HB-MoE layers; \textit{w/o Reg} uses the full architecture with $\lambda_{\text{AS}}{=}\lambda_{\text{HB}}{=}0$; and \textit{Single-MoE+Reg} applies both regularizers to a single non-hierarchical MoE layer. These variants separate the effects of sparse capacity, boundary action-space routing, heterogeneity balancing, and auxiliary regularization.

\subsection{Full Per-Step Results of HiMoE Component Ablations}
\label{appendix:hb-moe-full}

Table~\ref{tab:hb-moe-full} reports the complete per-sequence-length success counts (lengths 1--5) for the HiMoE component ablation on CALVIN ABC + D co-training. The aggregated \textbf{Sum.}\ column is also shown in the main paper (Table~\ref{tab:hb-moe}).

\begin{table}[ht]
    \centering
    \paperTable
    \caption{Full per-step results of the HiMoE component ablation on CALVIN ABC + D co-training. Each column 1--5 reports the average completion rate at the corresponding sequence length; \textbf{Sum.}\ is the cumulative count over the 5-step sequence.}
    \label{tab:hb-moe-full}
    \begin{tabular}{@{}c*{6}{c}@{}}
        \toprule
        \textbf{Setting} & \textbf{1} & \textbf{2} & \textbf{3} & \textbf{4} & \textbf{5} & \textbf{Sum.} \\
        \midrule
        w/o MoE        & \underline{0.918} & 0.837 & 0.744 & 0.681 & 0.597 & 3.777 \\
        Full-HB-MoE    & 0.917 & \underline{0.847} & \underline{0.774} & 0.713 & 0.650 & \underline{3.901} \\
        w/o AS-MoE     & 0.909 & 0.831 & 0.769 & \underline{0.718} & 0.646 & 3.873 \\
        w/o HB-MoE     & 0.904 & 0.826 & 0.749 & 0.708 & 0.649 & 3.836 \\
        w/o Reg        & 0.904 & 0.822 & 0.753 & 0.702 & \underline{0.654} & 3.835 \\
        Single-MoE+Reg & 0.914 & 0.839 & 0.757 & 0.688 & 0.615 & 3.813 \\
        Full           & \bf0.943 & \bf0.864 & \bf0.797 & \bf0.734 & \bf0.674 & \bf4.012 \\
        \bottomrule
    \end{tabular}
\end{table}

\subsection{Impact of Parameter Count vs. Architecture Design}
\label{appendix:param_efficiency}

In our main experiments (Table~\ref{tab:hb-moe}), the ``Full'' MoE setting utilizes $N=32$ experts with a top-$K$ routing of $K=4$. We note that this configuration results in slightly more active parameters compared to the ``w/o MoE'' baseline. To verify that the observed performance gains stem from the proposed mixture-of-experts design rather than merely increased model capacity, we conducted a control experiment with a parameter-matched dense baseline.

Specifically, we increased the parameter count of the ``w/o MoE'' model by enlarging the token feature dimensions and the FFN intermediate size, denoted as ``w/o MoE (Large)''. We co-trained this scaled baseline on CALVIN ABC (EEF) and CALVIN D (joint-angle) from scratch. As shown in Table~\ref{tab:param_scaling}, although increasing the parameters of the dense baseline (from 3.2B to 4.1B) yields a marginal improvement (Sum: 3.777 $\to$ 3.801), it still significantly lags behind our Full model (Sum: 4.012). Notably, our Full model achieves this superior performance with fewer active parameters (3.36B) than the scaled dense baseline (4.10B), confirming the effectiveness of the MoE architecture.

\begin{table}[ht]
    \centering
    \paperTable
    \caption{Comparison with a parameter-matched dense baseline. We scale up the ``w/o MoE'' baseline to match the total parameter count of our Full model. ``Param (All $|$ Act.)'' denotes total and active parameters in billions.}
    \label{tab:param_scaling}
    \begin{tabular}{@{}cc*{6}{c}@{}}
        \toprule
        \textbf{Method} & \textbf{Param (All $|$ Act.)} & \textbf{1} & \textbf{2} & \textbf{3} & \textbf{4} & \textbf{5} & \textbf{Sum.} \\
        \midrule
        w/o MoE & 3.24B $|$ 3.24B & 0.918 & 0.837 & 0.744 & 0.681 & 0.597 & 3.777 \\
        w/o MoE (Large) & 4.10B $|$ 4.10B & 0.898 & 0.806 & 0.744 & 0.702 & 0.651 & 3.801 \\
        Full (Ours) & 4.07B $|$ \textbf{3.36B} & \textbf{0.943} & \textbf{0.864} & \textbf{0.797} & \textbf{0.734} & \textbf{0.674} & \textbf{4.012} \\
        \bottomrule
    \end{tabular}
\end{table}

\subsection{MoE Warm-up}
\label{warmup}

The term ``MoE Warm-up'' refers to a two-stage fine-tuning strategy that prepares the MoE-related parameters before jointly optimizing the whole model. Although our goal is for HiMoE to abstract features into shared knowledge during pre-training, the domain gap between the pre-training mixture and a target fine-tuning dataset can be substantial. In such cases, directly updating all parameters from the first step forces the pre-trained backbone to absorb gradients from MoE routing and expert modules that have not yet adapted to the target domain, which can destabilize optimization.

Concretely, our warm-up proceeds in two stages. In the \emph{first stage}, we freeze all non-MoE parameters in the action module (i.e., the Transformer blocks that were pre-trained alongside and learned to cooperate with the MoE) as well as the VLM backbone, and update only the MoE-related parameters (routers, experts, and the shared expert) for a short warm-up phase. This stage lets the experts and routing distributions adapt to the target domain's action spaces and embodiments while preserving the pre-trained backbone. In the \emph{second stage}, we unfreeze all parameters and jointly fine-tune the entire model under the standard training objective. Thus, our evaluation measures fine-tuning transfer from heterogeneous pre-training.

\subsection{Analysis of Heterogeneous Co-training Pipeline}
\label{appendix:training_pipeline}
\textbf{Mechanisms for Heterogeneous Data.}
To effectively handle heterogeneous observation and action spaces during co-training, our pipeline incorporates two key mechanisms: \textbf{Data Mask} and \textbf{Loss Mask}.
\begin{itemize}
    \item \textbf{Data Mask:} To address varying state vector dimensions across tasks, we concatenate the state vector with a boolean mask. This mask explicitly indicates which indices correspond to valid data versus padding, allowing the model to distinguish between informative features and zero-padded values.
    \item \textbf{Loss Mask:} Instead of computing the loss over the full action vector (which may contain padded dimensions for certain datasets), we utilize a loss mask to compute the gradient only over the valid action dimensions specific to each task.
\end{itemize}

\textbf{Ablation Study.}
We validate the contribution of these components under the ``w/o MoE'' (Dense) setting with from-scratch co-training on CALVIN ABC + D. As shown in Table~\ref{tab:ablation_masks}, the full base setting achieves a total score of 3.777. Removing the \textit{Loss Mask} results in a performance drop to 3.696, indicating the importance of precise gradient supervision. Furthermore, removing both the \textit{Data Mask} and \textit{Loss Mask} leads to a significant degradation to 3.626. These results demonstrate that explicitly handling heterogeneity via masking strategies is critical for robust co-training performance.

\begin{table}[ht]
    \centering
    \paperTable
    \caption{Ablation of training pipeline components for heterogeneous co-training (CALVIN ABC + D). We evaluate the impact of Data Mask and Loss Mask on the dense (w/o MoE) baseline.}
    \label{tab:ablation_masks}
    \begin{tabular}{@{}c*{6}{c}@{}}
        \toprule
        \textbf{Method} & \textbf{1} & \textbf{2} & \textbf{3} & \textbf{4} & \textbf{5} & \textbf{Sum.} \\
        \midrule
        Base setting (with both masks) & \textbf{0.918} & \textbf{0.837} & \textbf{0.744} & \textbf{0.681} & 0.597 & \textbf{3.777} \\
        w/o Loss Mask & 0.896 & 0.800 & 0.726 & 0.667 & \textbf{0.607} & 3.696 \\
        w/o Data \& Loss Mask & 0.886 & 0.797 & 0.714 & 0.649 & 0.580 & 3.626 \\
        \bottomrule
    \end{tabular}
\end{table}

\subsection{State-Action Injection Ablation}
\label{appendix:state_injection}

For completeness, we evaluate whether the cross-action improvements depend on the specific way proprioceptive state information is injected into the action expert. We compare three state-injection mechanisms under the same from-scratch CALVIN protocols: (i) \textit{Concatenation}, our default design, which places state, noisy-action, and time tokens in a single Transformer sequence; (ii) \textit{Cross-Attention}, where action hidden states attend to state tokens; and (iii) \textit{AdaLN}, where state tokens are mapped to scale and shift parameters that modulate the action hidden states. Since changing the injection topology changes parameter shapes and prevents direct reuse of pre-trained weights, all variants are trained from scratch for a controlled comparison.

\begin{table}[ht]
    \centering
    \paperTable
    \caption{State-action injection ablation on CALVIN. We compare D-only training with CALVIN ABC (EEF) + D (joint-angle) co-training under the same HiMoE architecture.}
    \label{tab:state_injection}
    \begin{tabular}{@{}cccc@{}}
        \toprule
        \textbf{State-Injection Method} & \textbf{D (Joint) Only} & \textbf{ABC (EEF) + D (Joint)} & \textbf{Transfer Gain} \\
        \midrule
        Cross-Attention & 3.724 & 3.915 & $+0.191$ \\
        AdaLN           & 3.822 & 3.995 & $+0.173$ \\
        Concatenation   & 3.826 & \textbf{4.012} & $+0.186$ \\
        \bottomrule
    \end{tabular}
\end{table}

Table~\ref{tab:state_injection} shows that all three injection mechanisms obtain positive transfer when moving from single-action-space training to mixed EEF/joint co-training. This suggests that HiMoE's cross-action benefit is not an artifact of a particular state-injection interface. Concatenation gives the best mixed-action performance, supporting its use as a simple default for modeling bidirectional state-action correlations in a shared Transformer sequence.

\subsection{Analysis of Expert Routing}
\begin{figure*}[t]
    \centering
    \includegraphics[width=1\linewidth]{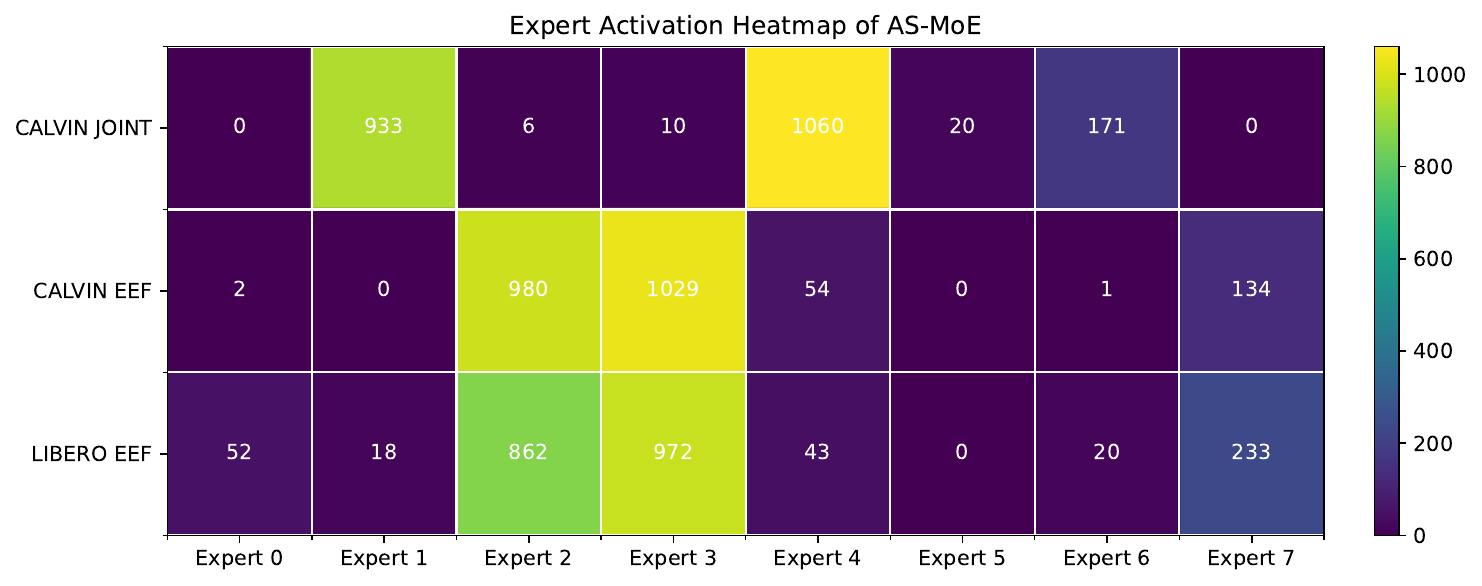}
    \caption{AS-MoE expert activation heatmap for the qualitative $N{=}8,K{=}2$ routing analysis on CALVIN joint-angle, CALVIN EEF, and LIBERO EEF.}
    \label{fig:asmoe-heatmap}
\end{figure*}

\begin{figure*}[t]
    \centering
    \includegraphics[width=1\linewidth]{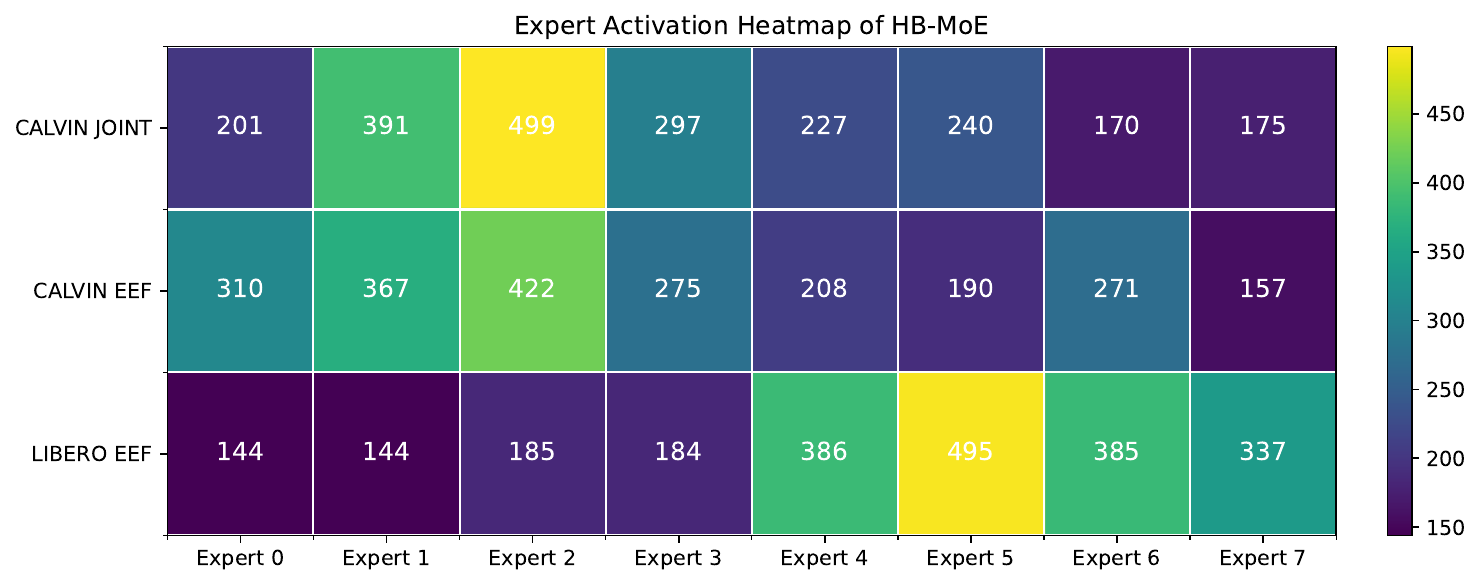}
    \caption{HB-MoE expert activation heatmap for the qualitative $N{=}8,K{=}2$ routing analysis on CALVIN joint-angle, CALVIN EEF, and LIBERO EEF.}
    \label{fig:hbmoe-heatmap}
\end{figure*}

For visualization purposes, we deliberately analyze a smaller HiMoE configuration with $N{=}8$ experts and top-$K{=}2$ routing on CALVIN joint-angle, CALVIN EEF, and LIBERO EEF, rather than the deployed $N{=}32, K{=}4$ setting. We stress that the figures in this subsection are intended as a \emph{qualitative illustration of the regularization mechanism}, not as deployed-scale performance evidence: the two patterns we wish to convey---(a) per-source clustering of the AS-MoE routing distribution and (b) approximately uniform expert utilization in HB-MoE---are directly encouraged by AS-Reg and HB-Reg. The choice of small $N{=}8,K{=}2$ is motivated purely by interpretability: (i) a $3{\times}32$ heatmap with top-4 routing is visually saturated and qualitatively indistinguishable across rows, while a $3{\times}8$ heatmap with top-2 routing makes the per-dataset specialization pattern visually unambiguous; (ii) a sparser top-$K$ ($K{=}2$ vs.\ $K{=}4$) yields a more peaked activation distribution that directly reveals which experts a given action space relies on, whereas $K{=}4$ averages out the contrast; (iii) at the deployed scale, expert specialization is partitioned across a much larger pool, so any single-panel color scale necessarily compresses the per-expert signal and obscures the very pattern we wish to visualize. The auxiliary losses are rescaled to be $N$-invariant (we use the DeepSeek-style $f_i{=}(N/KU)\sum_u r_{i,u}$ normalization with $\sum_i f_i {=} N$, and the InfoNCE-style contrastive ratio in AS-Reg is dimensionless in $N$). We further monitor both auxiliary losses during training at the deployed $\{N{=}32,K{=}4\}$ configuration and observe that HB-Reg converges close to its theoretical lower bound (corresponding to near-uniform expert utilization) and AS-Reg saturates at a similarly low value, suggesting similar routing behavior at deployed scale. We therefore use the $\{N{=}8,K{=}2\}$ panel purely as a diagnostic instrument for the routing mechanism; the downstream effect of this mechanism on task performance at the deployed $\{N{=}32,K{=}4\}$ scale is reported separately in Tables~\ref{tab:hetero-cotraining},~\ref{tab:hb-moe},~\ref{tab:sensor-hetero}, and~\ref{tab:experts-topk}. The expert activation heatmaps for AS-MoE and HB-MoE are shown in Figure~\ref{fig:asmoe-heatmap} and Figure~\ref{fig:hbmoe-heatmap}, respectively.

From Figure~\ref{fig:asmoe-heatmap}, we observe that CALVIN EEF and LIBERO EEF exhibit similar expert activation patterns, while CALVIN joint-angle shows a clearly different distribution, reflecting the differences in action space.
From Figure~\ref{fig:hbmoe-heatmap}, the expert activation patterns for CALVIN joint-angle and CALVIN EEF are similar, as these datasets share the same environment and observation settings except for the action space, whereas LIBERO EEF has a distinct activation pattern.

These visualizations provide qualitative evidence that the hierarchy encourages different routing behavior for action-space and observation-related variation.

\subsection{Cross-Action Space Fine-Tuning}
\label{appendix:cross-action-ft}

Beyond the from-scratch co-training studies in Tables~\ref{tab:hetero-cotraining} and~\ref{tab:hb-moe}, we further evaluate whether HiMoE can leverage heterogeneous action spaces during the \emph{fine-tuning} stage. Unlike the from-scratch setting, this experiment starts from pre-trained checkpoints and compares \emph{isolated} fine-tuning on a single action space against \emph{mixed} fine-tuning on both action spaces simultaneously. To make the comparison strict and fully reproducible, we use CALVIN, which supports the same task family under two distinct action parameterizations: ABC (EEF) and D (joint-angle).

\begin{table}[ht]
    \centering
    \paperTable
    \caption{Cross-action space fine-tuning on CALVIN. We fine-tune the pre-trained HiMoE-VLA either on a single action-space split (ABC (EEF) or D (joint-angle)) in isolation, or jointly on both splits. Numbers report the CALVIN long-horizon score (sum over 5 chained tasks); values in parentheses are the absolute gain over the corresponding isolated setting.}
    \label{tab:cross-action-ft}
    \begin{tabular}{@{}cccc@{}}
        \toprule
        \textbf{Evaluation setting} & \textbf{ABC (EEF) only} & \textbf{D (joint-angle) only} & \textbf{ABC (EEF) + D (joint-angle)} \\
        \midrule
        EEF evaluation   & 4.119 & --    & \bf 4.224 (\textbf{+0.105}) \\
        Joint-angle evaluation & --    & 3.967 & \bf 4.205 (\textbf{+0.238}) \\
        \bottomrule
    \end{tabular}
\end{table}

As shown in Table~\ref{tab:cross-action-ft}, mixed fine-tuning consistently outperforms isolated fine-tuning on both the EEF ($+0.105$) and joint-angle ($+0.238$) action spaces, with a particularly large gain on the joint-angle split where data is harder to learn from scratch. Moreover, both mixed-tuning scores are substantially higher than the corresponding from-scratch results in Table~\ref{tab:hetero-cotraining}, indicating that the pre-trained HiMoE provides transferable physical priors that are further amplified when heterogeneous action data is mixed during fine-tuning. Together with the shared-action observation/scene co-training experiment in Table~\ref{tab:sensor-hetero}, these results support the effectiveness of the hierarchy under both action-space and non-action heterogeneity rather than merely benefiting from extra capacity.

We further compare against a dense $\pi_0$-style baseline under the same CALVIN D joint-angle evaluation. As shown in Table~\ref{tab:cross-action-ft-baseline}, mixed fine-tuning provides only a modest gain for the dense baseline, whereas HiMoE obtains a larger improvement and a higher absolute score. This suggests that the benefit of mixed-action data is substantially amplified by the proposed hierarchical MoE action module.

\begin{table}[ht]
    \centering
    \paperTable
    \caption{Dense baseline vs. HiMoE under CALVIN mixed-action fine-tuning from pre-trained checkpoints.}
    \label{tab:cross-action-ft-baseline}
    \begin{tabular}{@{}cccc@{}}
        \toprule
        \textbf{Method} & \textbf{D (Joint) Only} & \textbf{ABC (EEF) + D (Joint)} & \textbf{Transfer Gain} \\
        \midrule
        Dense baseline & 3.756 & 3.816 & $+0.060$ \\
        HiMoE-VLA      & 3.967 & \textbf{4.205} & $+0.238$ \\
        \bottomrule
    \end{tabular}
\end{table}

\subsection{Training and Inference Efficiency}
\label{appendix:efficiency}

We benchmark the training and inference cost of HiMoE-VLA to quantify the overhead introduced by the hierarchical MoE design.

\textbf{Training speed.}
Training throughput is measured on the CALVIN benchmark using 8$\times$NVIDIA A100 (40\,GB) GPUs with a global batch size of 32, DeepSpeed ZeRO-Stage~2, and gradient checkpointing. As shown in Table~\ref{tab:train-speed}, the full HiMoE configuration ($N{=}32$, $K{=}4$) incurs only $\sim$7\% additional per-iteration cost relative to the dense (\textit{w/o MoE}) variant, while delivering substantially better performance on heterogeneous data (Tables~\ref{tab:hetero-cotraining},~\ref{tab:hb-moe}).

\begin{table}[ht]
    \centering
    \paperTable
    \caption{Training speed (seconds per iteration) on CALVIN with 8$\times$NVIDIA A100 (40\,GB) GPUs, a global batch size of 32, DeepSpeed ZeRO-Stage~2, and gradient checkpointing.}
    \label{tab:train-speed}
    \begin{tabular}{@{}ccccc@{}}
        \toprule
        \textbf{Model} & w/o MoE & $N{=}8,K{=}2$ & $N{=}8,K{=}4$ & $N{=}32,K{=}4$ \\
        \midrule
        Speed (s/iter) & 1.14 & 1.18 & 1.20 & 1.22 \\
        \bottomrule
    \end{tabular}
\end{table}

\textbf{Inference speed.}
Inference latency is measured on a single NVIDIA RTX 4090 GPU. We note that $\pi_0$ benefits from a highly optimized JAX implementation, whereas all other evaluated baselines and our method are implemented in PyTorch. As reported in Table~\ref{tab:infer-speed}, introducing the MoE routing mechanism inevitably incurs a slight increase in inference latency, yet our overall inference speed remains highly competitive with existing mainstream efficient methods such as OpenVLA-OFT and RDT-1B.

\begin{table}[ht]
    \centering
    \paperTableTight
    \caption{Inference latency (seconds per action chunk) on a single NVIDIA RTX 4090 GPU. $\pi_0$ uses a highly optimized JAX implementation; all other methods (including ours) are implemented in PyTorch.}
    \label{tab:infer-speed}
    \begin{tabular}{@{}c*{8}{c}@{}}
        \toprule
        \textbf{Model} & \multicolumn{4}{c}{\textbf{Baselines}} & \multicolumn{4}{c}{\textbf{Ours}} \\
        \cmidrule(lr){2-5}\cmidrule(l){6-9}
        & OpenVLA & \mbox{OpenVLA-OFT} & RDT-1B & $\pi_0$ & w/o MoE & $N{=}8,K{=}2$ & $N{=}8,K{=}4$ & $N{=}32,K{=}4$ \\
        \midrule
        Lat.\ (s) & 0.307 & 0.182 & 0.168 & 0.086 & 0.153 & 0.185 & 0.193 & 0.195 \\
        \bottomrule
    \end{tabular}
\end{table}


\section{More Visualizations}
\begin{figure*}[t]
    \centering
    \includegraphics[width=1\linewidth]{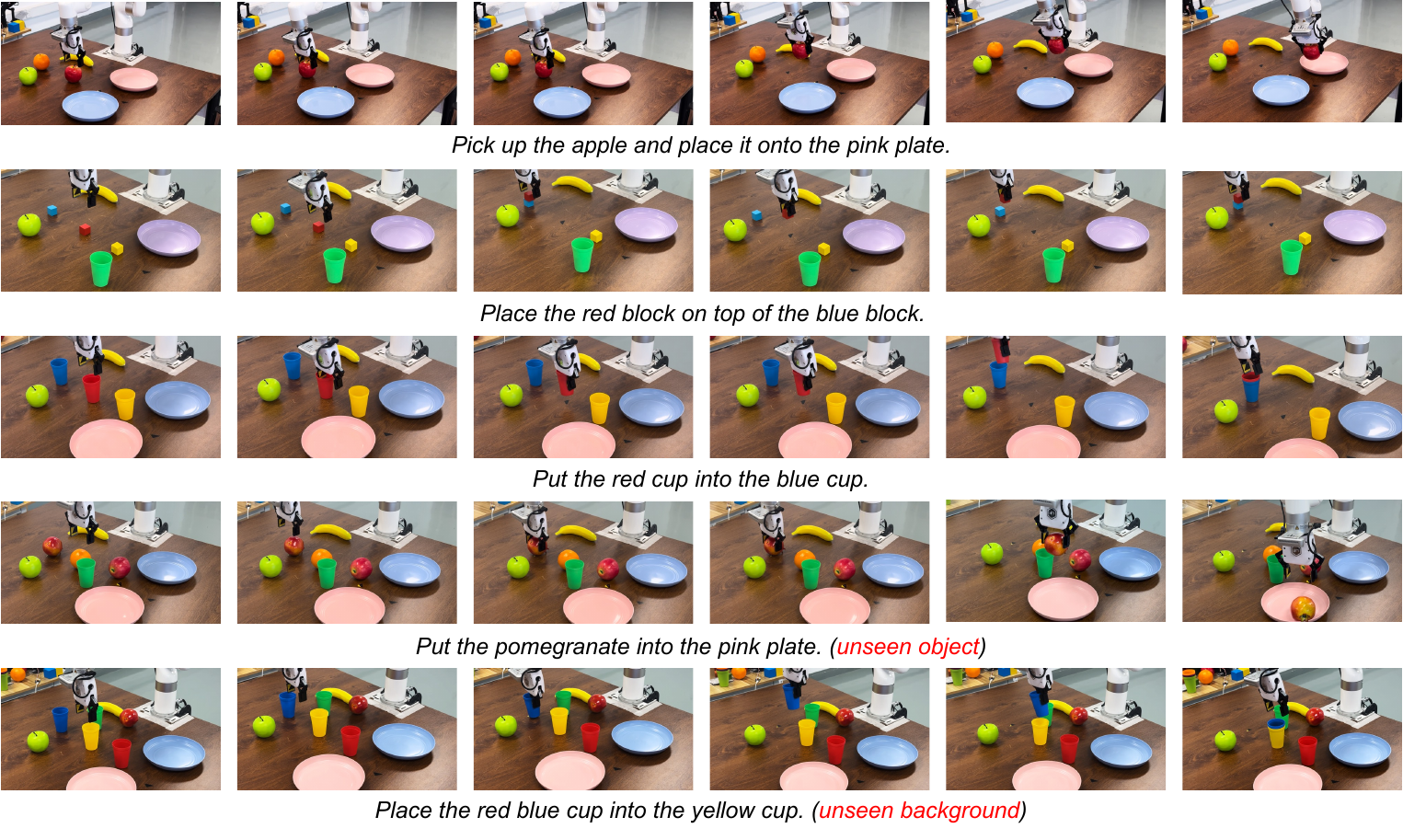}
    \caption{Overview of tasks on the single-arm xArm7 robot, including seen tasks and generalization settings with unseen distractors or novel objects.}
    \label{fig:xarm2}
\end{figure*}

\begin{figure*}[t]
    \centering
    \includegraphics[width=1\linewidth]{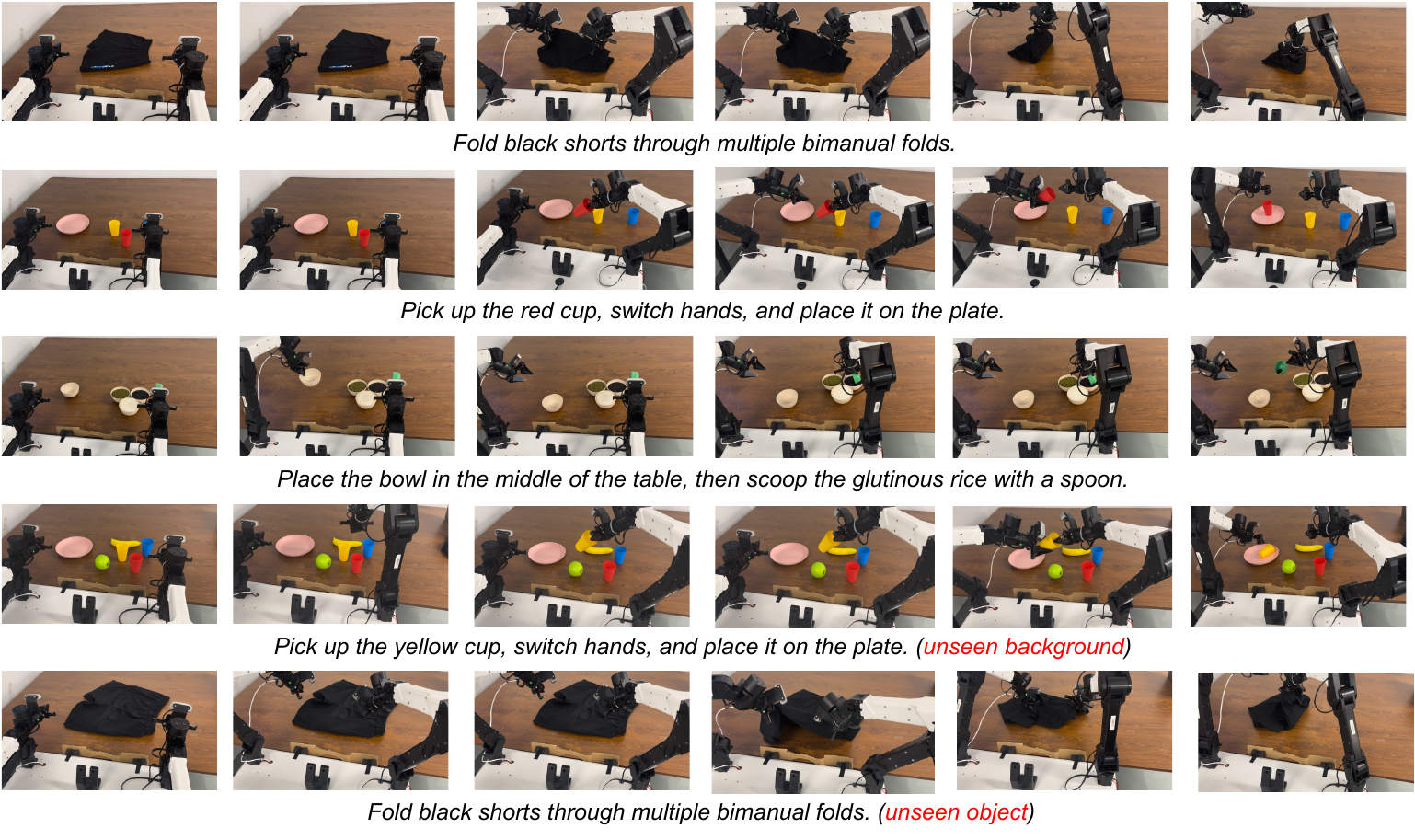}
    \caption{Overview of tasks on the dual-arm ALOHA robot, including seen tasks and generalization settings with unseen distractors or a novel garment.}
    \label{fig:aloha}
\end{figure*}

Additional task-level visualizations are shown in Fig.~\ref{fig:xarm2} and Fig.~\ref{fig:aloha}.

\end{document}